\documentclass[journal,transmag]{IEEEtran}
\usepackage{graphicx}
\usepackage{amssymb}
\usepackage{subfigure}
\usepackage{makecell}
\usepackage{color}
\usepackage{multirow}
\usepackage{rotating}
\usepackage{adjustbox}
\usepackage{tabularx}
\ifCLASSINFOpdf

\else

\fi
\usepackage{xcolor}

\usepackage{amsmath}
\usepackage{mathtools,nccmath}

\usepackage[linesnumbered,boxed,ruled,commentsnumbered]{algorithm2e}

\SetCommentSty{mycommfont}

\begin{document}

\title{
DySR: A Dynamic Representation Learning and Aligning based Model for Service Bundle Recommendation
}

\author{\IEEEauthorblockN{Mingyi Liu,
Zhiying Tu~\IEEEmembership{Member,~IEEE},
Xiaofei Xu~\IEEEmembership{Member,~IEEE}, 
Zhongjie Wang~\IEEEmembership{Member,~IEEE}}
\thanks{Mingyi Liu, Zhiying Tu, Xiaofei Xu, and Zhongjie Wang are with the Faculty of Computing, Harbin Institute of Technology, Harbin, China (e-mail: liumy@hit.edu.cn, tzy\_hit@hit.edu.cn, xiaofei@hit.edu.cn, rainy@hit.edu.cn)}

\thanks{Manuscript received XXXXXX; XXXXXXXXX. 
Corresponding author: Zhongjie Wang (email: rainy@hit.edu.cn).}

}



\markboth{Journal of \LaTeX\ Class Files,~Vol.~14, No.~8, August~2015}%
{Shell \MakeLowercase{\textit{et al.}}: Bare Demo of IEEEtran.cls for IEEE Transactions on Magnetics Journals}

\IEEEtitleabstractindextext{%
\begin{abstract}
An increasing number and diversity of services are available, which result in significant challenges to effective reuse service during requirement satisfaction. There have been many service bundle recommendation studies and achieved remarkable results. However, there is still plenty of room for improvement in the performance of these methods. The fundamental problem with these studies is that they ignore the evolution of services over time and the representation gap between services and requirements. In this paper, we propose a dynamic representation learning and aligning based model called DySR to tackle these issues. DySR eliminates the representation gap between services and requirements by learning a transformation function and obtains service representations in an evolving social 
environment through dynamic graph representation learning. Extensive experiments conducted on a real-world dataset from ProgrammableWeb show that DySR outperforms existing state-of-the-art methods in commonly used evaluation metrics, improving $F1@5$ from $36.1\%$ to $69.3\%$.

\end{abstract}

}

\maketitle

\IEEEdisplaynontitleabstractindextext

\IEEEpeerreviewmaketitle

\section{Introduction}\label{sec:intro}
With the rapid development of new technologies such as cloud, edge, and mobile computing, the number and diversity of available services are dramatically exploding, and services have become increasingly important to people's daily work and life. The increasing number and diversity of services bring significant challenges to effective service management and reuse. Consequently, recommending suitable services for developers based on requirements to help them reduce the burden of service selection in creating applications (mashups) becomes a non-trivial issue, and such task is often called service bundle recommendation in services computing.

Service bundle recommendation refers to recommending a set of services based on these services' functions and history compositions to satisfy the user's explicit and implicit requirements\cite{WU2015133}. There have been many service recommendation studies from different perspectives. Existing approaches can be categorized into natural language processing (NLP) based approaches\cite{faieq2019context,lei2015time,lin2018nl2api}, graph-based approaches\cite{hu2019poi,chang2021graph,cao2019efficient}, and hybrid approaches \cite{gao2016seco,ma2021deep}. These approaches have achieved remarkable results. However, ignoring the following factors  limits the performance of their approaches and the application scenarios in reality:

    (1) \textbf{Cold Start:}  Most existing approaches leverage the interaction history between mashups and services to recommend missing services. They work well in normal situations like user-item recommendations and user-movie recommendations. The co-occurrence between users(mashups) and items(movies/services) can be repeated. However, in the application scenario of mashup creation, each mashup only appears once and without any interaction with existing services. The lack of such information decreases the performance of existing approaches. This phenomenon is also known as the \textbf{cold start of requirements}, which has recently started to attract the attention of researchers\cite{ma2021deep,gu2021csbr}. Additionally, existing approaches assume all services have enough description document and historical usage information (e.g., invocation history, co-invocation, and popularity). While, in the real world, services are created in temporal order, and newly created services often do not accumulate enough historical usage information or even no historical usage information. Or, due to unregulated development, some service providers do not provide a comprehensive service description. These services are usually ignored in existing approaches. We call this phenomenon the \textbf{cold start of services}, which has not received enough attention in service bundle recommendations.
    
    (2) \textbf{Evolving Service Social:} Since, in most cases, services do not actively update their profile promptly\cite{wang2021external}, existing approaches assume services' function and quality are static. However, we argue that the quality of services evolves, and the functionality of services evolves in a latent way as their social environment changes. For example, a service that other applications have not invoked for a long time may indicate that the service quality has degraded or that an alternative service with better similar functionality is available and should be avoided when making service recommendations.  
    \textit{Readmill}\footnote{https://www.programmableweb.com/api/readmill} is an example of service function evolving\cite{bai2017sr}, which is designed for ``Social'' as its profile says ``\textit{Readmill is an online and mobile platform for readers to share information about ..., allowing them to highlight and discuss sections of eBooks with other users. . .}''. The mashups that invoke \textit{Readmill} are gradually moving from the ``Social'' domain to the ``Book'' domain, indicating that the social environment of \textit{Readmill} is gradually evolving from ``Social'' to ``Books''. 
    
   (3) \textbf{Representation Gaps between Services and Requirements:} There is a large gap between services and requirements representation, often overlooked in existing approaches. The representation of requirements focuses on the functionality and value that the constructed mashup can provide to the user and is accompanied by an extensive domain vocabulary. 
    In contrast, the representation of services is more freedom. Services can be described using natural language text, but they may describe their performance and input/output format in a technical-oriented language style. Services can also be represented using a set of properties, such as QoS, popularity. 
    For example, \textit{QuickMocker}\footnote{https://www.programmableweb.com/mashup/quickmocker} is a mashup where the requirement it satisfied is described as \textit{``QuickMocker is an online API tool that allows to create your own public domain and a fake web services...''}. While its component service \textit{Mocky}\footnote{https://www.programmableweb.com/api/mocky} are described as \textit{``Mocky is available as a web console, downloadable software, or REST API. The API version of the service accepts calls issued in JSON and JSONP. The Mocky website is available in English, French, and Portuguese''}.
    These representation gaps make services and requirements exist in different semantic spaces. The arbitrary mixing of features in two different spaces in existing methods can significantly affect the performance of recommendations.

\begin{figure}[!h]
    \centering
    \includegraphics[width=\linewidth]{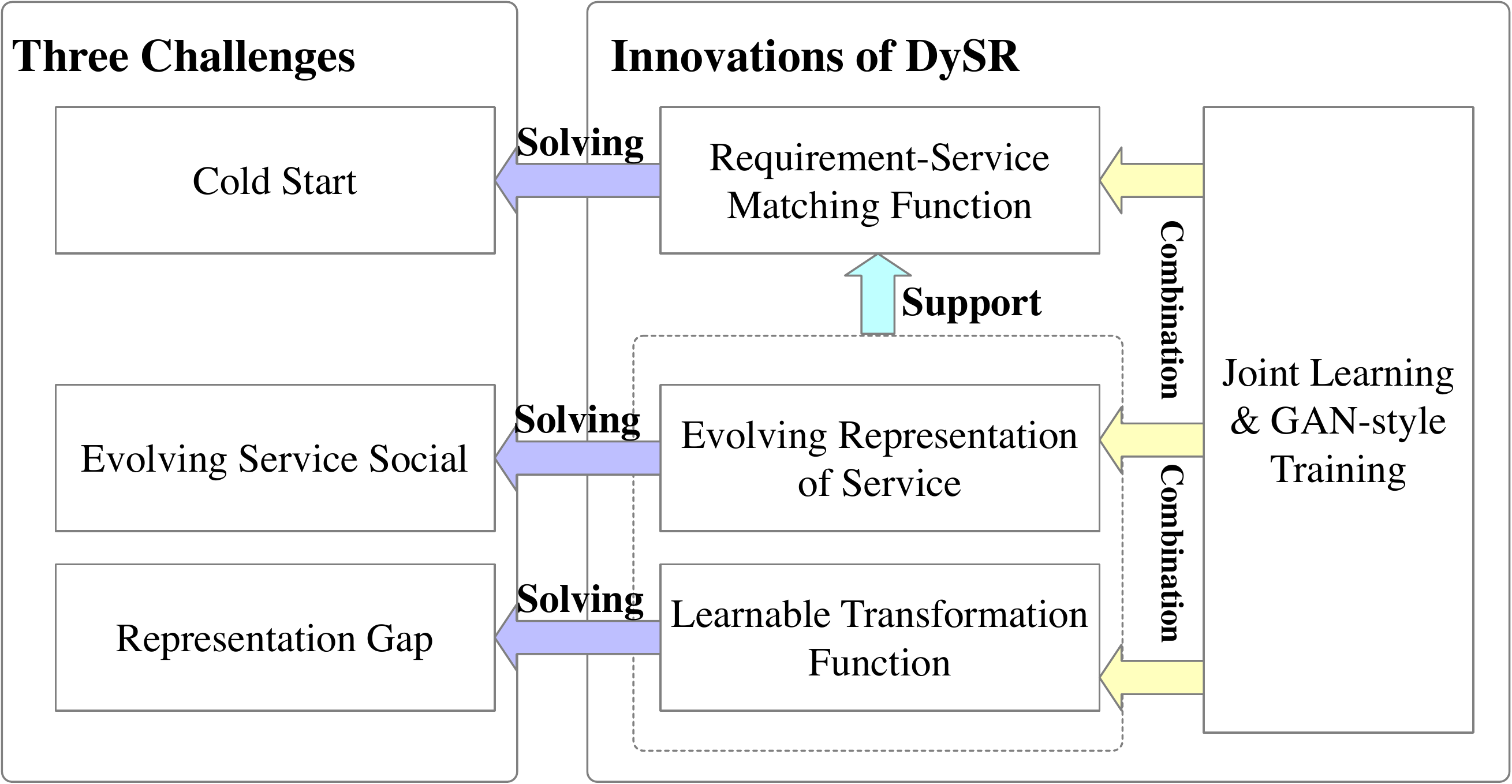}
    \caption{The correspondence between the innovations of our proposed DySR model and the service bundle recommendation challenges}
    \label{fig:mapping}
\end{figure}
This paper proposes a dynamic representation learning and aligning based model called DySR to solve the three challenges mentioned above. Fig.~\ref{fig:mapping} shows how the innovation modules in DySR address the challenges mentioned above, where \textbf{Learnable Transformation Function}, \textbf{Evolving Representation of Service}, and \textbf{Requirement-Service Matching Function} are used to address \textbf{Representation Gaps}, \textbf{Evolving Service Social}, and \textbf{Cold Start}, respectively. In aadition, \textbf{Joint Learning \& GAN-style Training} is used to combine the other three modules so that they can provide complementary information to each other:
\begin{enumerate}
    \item \textbf{Learnable Transformation Function}: We propose a learnable transformation function that aligns the representation of requirements to the semantic space where the representation of services resides, thus eliminating the representation gap between requirements and services.
    \item \textbf{Evolving Representation of Service}: We form a dynamically evolving graph of services based on co-invocation relations. Then we perform dynamic representation learning on this dynamic graph to obtain a dynamic representation of the services to solve evolving service social environment.
    \item \textbf{Requirement-Service Matching Function}: We introduce a learnable function called the requirement-service matching function to evaluate the probability of an invocation between a given requirement and a service. This function works with only the aligned representation of the requirements and services and does not require other information (e.g., known component service, historical usage). Thus the cold start of mashups and the cold start of services are solved at the same time.
    \item \textbf{Joint Learning \& GAN-style Training}: DySR models the influence mechanism of service and requirement by jointly learning the dynamic representation learning task of services and service recommendation task. A Generative Adversarial Network (GAN)\cite{2014Generative} like training approach is also proposed to model the process of requirement-induced service evolution and service evolution reacting to requirement, unlike GAN, where the two tasks are not adversarial but cooperative.
\end{enumerate}

Besides the proposed DySR model, the contributions of this paper also include: 1) We provide an in-depth analysis of the issues that have not been sufficiently discussed in real-world service recommendation scenarios; 2) We have conducted extensive experiments on the real-world dataset ProgrammableWeb, which shows that our proposed model significantly outperforms several state-of-the-art approaches in terms of recommendation performance; 3) We have also constructed additional experiments to discuss the effects of the different components of DySR.

The remainder of this paper is organized as follows: In Section \ref{sec:related_work}, we introduce related work. In Section \ref{sec:model}, we describe relevant details of the DySR model. In Section~\ref{sec:settings}, we give the details of the experiment settings. In Section~\ref{sec:results}, we present the experiment results and discuss how DySR works. In the final section, we present the conclusion.

\section{Related Work}\label{sec:related_work}
The use of service bundle recommendations to satisfy the requirements of users creating new mashups has recently attracted a great deal of academic and industrial attention. Service recommendation approaches can be divided into the following categories based on the information used: NLP-based approaches, graph-based approaches, and hybrid approaches.

\subsection{NLP-based Approaches}
The main idea of NLP-based approaches is to make recommendations by measuring the similarity between the description of the candidate services and the requirement. Keywords\cite{he2016keyword,he2017efficient} and TF-IDF\cite{xia2014category} are used to match services to satisfy mashup creation. However, these matching procedures cannot really understand semantic and suffer from poor performance.

Ontology-based approaches annotate requirements and services with domain ontologies and use these ontologies to match high-level concepts or calculate their semantic similarities\cite{al2015semantic,karthikeyan2018fuzzy,rupasingha2019alleviating}. However, the lack of suitable domain ontologies and the huge manual annotation cost of ontology construction make ontology-based approaches difficult to use in real-world scenarios. Some studies attempt to use latent semantics to denote text features. Currently, the most common technique used to solve this problem in the field of service computing is the topic model\cite{lin2018nl2api,gu2021csbr}. For example, Li et al.~\cite{li2014novel} use a topic model to explore the semantic relationships between mashups and services. Zhong et al.~\cite{zhong2016web} refactor descriptions by using Author-Topic Model~\cite{rosen2012author} to eliminate the gaps between mashups and services. However, small data volumes, high noise in data, and ignoring word orders make the performance of topic model-based methods unsatisfactory; for example, the recent SOTA method~\cite{gu2021csbr} based on topic models has an $F1@5$ value of only about \textbf{30\%} on the ProgrammableWeb dataset. With the great success of pre-trained language models (PLMs) and deep learning models in the NLP, some researchers start using PLMs and deep learning to solve the service recommendation problem. For example, Bai et al.~\cite{bai2017dltsr} designed a stacked denoising autoencoder (SADE) to extract features for the recommendation. However, experimental results show that the simple introduction of PLMs and deep learning does not significantly improve recommendation performance, possibly due to the inability of existing models to directly adapt to service recommendation and negative transfer~\cite{wang2019characterizing}.

The major drawback of the NLP-based approaches is the inability to leverage the social information between services, leading to recommended service bundles with combinations of services that cannot co-occur due to real-world constraints, despite being functionally compatible.

\subsection{Graph-based Approaches}
Unlike NLP-based approaches, graph-based approaches make recommendations by mining information from historical interaction between services or requirements (users). For example, \cite{wang2019duskg,hu2019poi,mezni2021context} design recommendation algorithms based on the feature learned on knowledge graph. 

The graph-based approach usually works in conjunction with collaborative filtering (CF). For example, Chen et al.~\cite{zheng2012collaborative,chen2010regionknn} propose a neighborhood integrated matrix factorization approach to predict the quality of service (QoS) of candidate services. Chang et al.~\cite{chang2021graph} designed a graph-based matrix factorization approach to predict QoS, then use the QoS to select services. Besides predicting QoS, graph is also applied to find similar users or services. For example, Maardji et al.~\cite{maaradji2011social} proposes a frequent pair mining method for mashup development. Qi et al.~\cite{qi2017data} adopts a hybrid random walk to compute the similarities between users or services, and a CF model is designed for service recommendation. \cite{xie2019integrated,liang2016meta} build a heterogeneous information network using various information of services and mashups to measure the similarity between mashups, and then use the user-based CF to ranking candidate services.

Graph-based approaches cannot solve the cold start problem, as new requirements do not have any historical information.

\subsection{Hybrid Approaches}
Considering the complementarity of textual and graph information, a number of hybrid service recommendation approaches combining the two types of data have been proposed in recent years.

Li et al.~\cite{li2014novel} add the invocation relations between requirements and services to a latent Dirichlet allocation (LDA) model to enable the topic model learn the relationship between services and requirements, . \cite{gao2016seco,xia2014category} incorporates data structure made up of service and their co-invocation records into LDA. Jain et al.~\cite{jain2015aggregating} and Samanta et al.~\cite{samanta2017recommending} use topic models and neighbor interaction probabilities to calculate similarity scores between services and requirements, then multiply these scores to rank candidate services.

Deep learning-based approaches are becoming the mainstream of hybrid methods. For example, Xiong et al.~\cite{xiong2018deep} integrates the invocation relations between services and requirements as well as their description similarity into a deep neural network (DNN). Chen et al.~\cite{chen2018software} propose a preference-based neural collaborative filtering\cite{he2017neural} recommendation model, which uses multi-layer perceptron to capture the non-linear user-item relationships and obtain abstract  data  representation from  sparse  vectors. Ma et al.~\cite{ma2021deep} utilize the powerful representation learning abilities provided by deep learning to extract textual features and features from various types of interactions between mashups and services. Wu et al.~\cite{9492754}  propose a neural framework based on multi-model fusion and multi-task learning, whiche exploits a semantic component to generate representations of requirements and introduces a feature interaction component to model the feature interaction between mashups and services.

Although the hybrid approach is a significant improvement over the NLP-based approaches and graph-based approaches, the current SOTA approach\cite{ma2021deep} still has an F1@5 value below \textbf{40\%} on the ProgrammableWeb dataset. This is mainly caused by ignoring the difference in representation between services and requirements and the evolving service social environment. The DySR model proposed in this paper is a hybrid approach that achieves $F1@5$ values close to \textbf{70\%} on the ProgrammableWeb dataset by tackling the representation gap and service evolution.

\section{Dynamic Representation Learning and Aligning based Model}\label{sec:model}
\begin{table}[htbp]
\centering
\caption{Notions used in this paper}\label{tab:notions}

\begin{tabularx}{0.9\linewidth}{cX}
\hline
Notion                   & \multicolumn{1}{c}{Description}                                                                           \\ \hline
$r$                      & Requirement                                                                                           \\
$s$                      & Service                                                                                               \\
$t$                      & Timestamp                                                                                             \\
$\mathbf{v}_r$           & The latent representation of requirement $r$                                                          \\
$Z^t$                    & The set of all services latent representation updated after time $t$                                             \\
$\mathbf{z}^t_s$         & The representation of service $s$ being updated after a co-invocation event involving $s$ at time $t$ \\
$\mathbf{z}^{\bar{t}}_s$ & Most recently updated latent representation of service $s$ just before $t$                            \\
$\Psi$                   & Transformation function used for align representation of requirements and services                    \\
$\Lambda$                & Requirement-Service matching Function                                                                   \\
$\hat{y}_{r,s}^t$          & The probability of $s$ being a component service of $r$ at time $t$
                            \\
$\Omega$                 & The set of trainable parameters in the unsupervised learning module                                   \\
$\mathcal{S}^t$          & Temporary attention matrix at time t                                                                  \\
$A^t$                    & Service-Service co-invocation adjacency matrix at time $t$      \\
$\lambda_{s_1, s_2}^t$   & Conditional intensity between $s_1$ and $s_2$ at time $t$                                             \\
$o$                      & $o=(s_1, s_2, t)$ denotes a co-invocation event involving $s_1$ and $s_2$ at time $t$                 \\ \hline
\end{tabularx}
\end{table}

In this section, we introduce the details of the DySR model. In Section~\ref{sec:framework} we first explain the joint learning framework of DySR. Next, we detail two main subtasks in DySR, supervised service recommendation task and unsupervised evolving service representation task,in Section~\ref{sec:supervised} and Section~\ref{sec:unsupervised}, respectively. Finally, in Section~\ref{sec:gan-style} we describe the GAN-style training process. Table~\ref{tab:notions} list frequently-used notions in this section and their meanings.

\subsection{Overall Framework}\label{sec:framework}
\begin{figure*}[tbp]
    \centering
    \includegraphics[width=\linewidth]{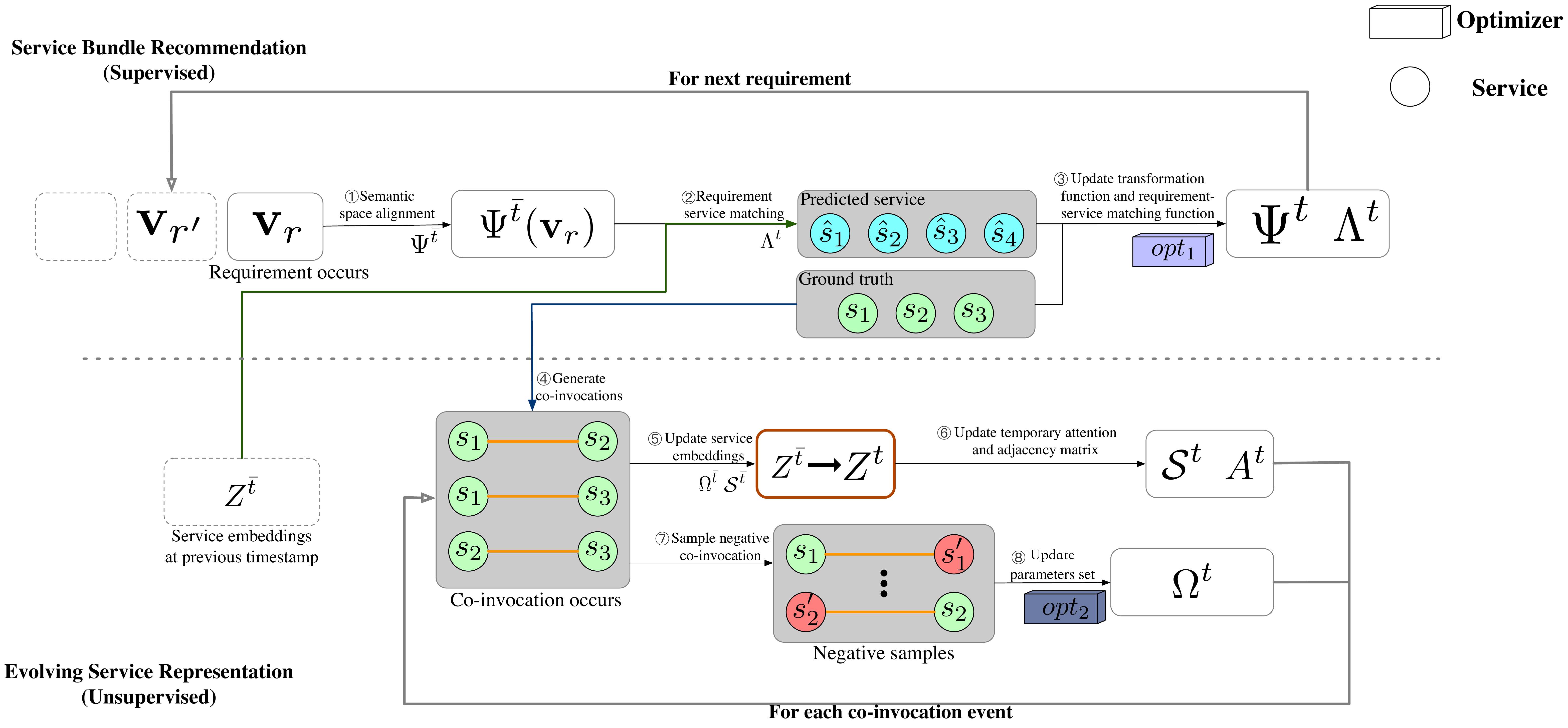}
    \caption{Overview framework of DySR model}
    \label{fig:framework}
\end{figure*}
As shown in Fig.~\ref{fig:framework}, the proposed DySR model consists of two subtasks: a supervised service bundle recommendation task and an unsupervised evolving service representation task. The service recommendation subtask is used to satisfy our need to recommend suitable services for a new requirement, and this task is used in both the training and inference stage of the DySR model. The representation gaps between requirement and service are eliminated, and the requirement-service matching function is learned in this subtask. The evolving service representation subtask is used to solve the service evolving problem. It is a supplementary task to the service recommendation subtask and is used to provide a suitable service representation for service bundle recommendation, so it will only be used in the training stage of the DySR model.

During the training stage, the input of DySR model is a requirement $r$ that is satisfied at historical time $t$ by a set of component services $C^+=\{s_1, s_2,\dots,s_n\}$, which is used in the supervised task as ground truth. And a set of co-invocation event $\mathcal{O}=\{o=(s_u, s_v, t) | s_u, s_v \in C^+ \text{ and } s_u \ne s_v\}$ is generated from $C^+$ as the training samples of the unsupervised task. While in the inference stage, the input of DySR model is a requirement $r$, and the output is a set of recommended component services $\hat{C}=\{\hat{s}_1, \hat{s}_2, \dots, \hat{s}_m\}$.

\subsection{Supervised Service Bundle Recommendation}\label{sec:supervised}
For a requirement $r$ at time $t$, we denote the latent representation as $\mathbf{v}_r \in \mathbb{R}^d_r$, a dense vector obtained by a certain encoding method (e.g. text encoding using pre-trained language model), where $d_r$ is the dimension of the requirement's latent representation. We use $\mathbf{z}^{\bar{t}}_{s} \in \mathbb{R}^{d_s}$ to denote the recently updated latent representation of  candidate service $s$ just before $t$, where $d_s$ is the dimension of each service's latent representation. And $Z^{\bar{t}}$ is all candidate services latent representation. The obtaining of $Z^{\bar{t}}$ is discussed in detail in Section~\ref{sec:unsupervised}.

\subsubsection{Transformation Function}
\begin{figure}[hbtp]
    \centering
    \includegraphics[width=\linewidth]{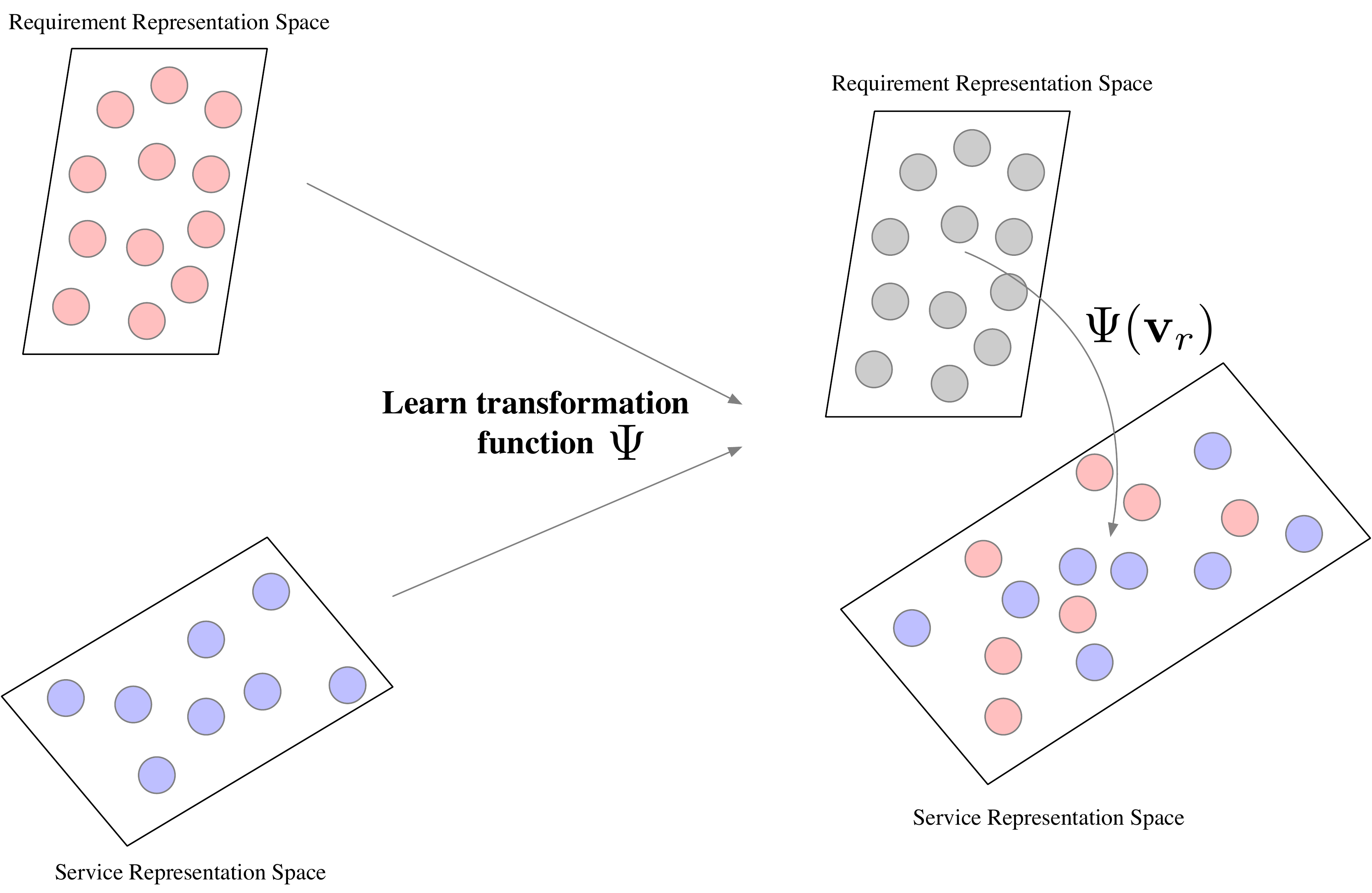}
    \caption{Illustration of transformation function}
    \label{fig:transformation}
\end{figure}
As we state in Section~\ref{sec:intro}, a key problem in real-world service recommendation is to eliminate the gap between requirement representations and service representations. Transformation function $\Psi$ can be learned to align the representation space to solve this problem, and the overall idea is illustrated in Fig.~\ref{fig:transformation}.

In this paper, we use affine transformation\cite{2016Affine} to implement the transformation function $\Psi$:
\begin{equation}\label{eq:transformation}
    \Psi (\mathbf{v}_r) = W_{\Psi} \centerdot \mathbf{v}_r + \mathbf{b}_{\Psi}
\end{equation}
where $W_{\Psi} \in \mathbb{R}^{d_r \times d_s}$ is affine transformation matrix and $\mathbf{b}_{\Psi} \in \mathbb{R}^{d_s}$ is translation vector. The main advantage of affine transformation is preserves collinearity and ratios of distance, which means that similar requirement remain similar after affine transformation.

\subsubsection{Requirement-Service Matching Function}
After transforming the requirement representation into the same semantic space of the service representations. We designed a requirement-service matching function that first fuses the representations of requirement $r$ and candidate service $s$ under the same semantic space and then outputs the probability that $s$ is a component service of $r$ at time $t$, $\hat{y}_{r,s}^t$:
\begin{equation}\label{eq:rs}
    \Lambda(\Psi (\mathbf{v}_r), \mathbf{z}_s^{\bar{t}}) =  \mathbf{z}_s^{\bar{t}} \centerdot W_{\Lambda} \centerdot \mathbf{v}_{r}^\intercal
\end{equation}
where $\intercal$ is vector transpose operation and $W_{\Lambda} \in \mathbb{R}^{d_s \times d_s}$ is a trainable parameter. Since we want to know the probability, we adopt a sigmoid activation function in the output the requirement-service matching function:
\begin{equation}
    \hat{y}^t_{r,s} = \sigma(\Lambda(\Psi (\mathbf{v}_r), \mathbf{z}_s^{\bar{t}}))
\end{equation}

\subsubsection{Loss Function}
For a given requirement $r$ with component services $C^+=\{s_1, s_2,\dots,s_n\}$ at time $t$, we minimize the following loss function:
\begin{equation}\label{eq:loss1}
    \mathcal{L}_{1} = -\sum_{s \in C^+ \cup C^-}y_{r,s}^t\log\hat{y}_{r,s}^t + (1-y_{r,s}^t)\log(1-\hat{y}_{r,s}^t)
\end{equation}
where $y_{r,s}^t \in \{0, 1\}$ denotes whether $s$ is a component service of $r$ at time $t$, and $C^-$ denotes a set of negative samples with services that are not component services of $r$ at time $t$. Usually, a requirement is satisfied using a limited number of services, but the number of candidate services is much larger than the number of services required. So it is not appropriate to use all unselected services as negative samples, and we select negative samples of number $|C^-|=6|C^+|$ by random sampling.

\subsection{Unsupervised Evolving Service Representation}\label{sec:unsupervised}
Unlike existing approaches that maintain a requirement-service invocation graph, DySR only needs to maintain a dynamic service co-invocation graph, which is more practical:
\begin{itemize}
    \item Compared to services, requirements are one-off and grows rapidly in volume, making the requirement-service invocation graph very large and sparse over time. On the one hand, this does not facilitate the extraction of information, and on the other hand, requires more expensive hardware resources.
    \item Requirements are usually created by users, which may involve privacy and security issues. For example, users may not want their requirements to be available to other users. Service co-invocation graph can solve this problem, as it only records co-invocations between services but not specific requirement content.
\end{itemize}
The service co-invocation graph can be represented as a dynamic adjacency matrix $A^t \in \mathbb{R}^{|\mathcal{C}^t| \times |\mathcal{C}^t|}$, where $\mathcal{C}^t$ is the set of services at time $t$, and $A^t_{s_u, s_v}$ is the number of times service $s_u$ and $s_v$ have been co-invoked before time $t$. The phenomenon of service evolution can be reflected by the evolution of the service co-invocation graph, which can be seen as a co-invocation event-driven temporal point process. We use $o=(s_u, s_v, t)$ to represent a co-invocation event, which means $s_u$ and $s_v$ are co-invoked at time $t$.

We use a variant of the DyRep\cite{trivedi2019dyrep} model to learn the service representation on a dynamic service co-invocation graph. For a given co-invocation event $o=(s_u, s_v, t) \in \mathcal{O}$, the main training steps consist of the following:

\subsubsection{Service Representation Update}
When an co-invocation event occurs, the representation of participating service $s_u$ is updated based on the three terms of \textbf{Self-propagation}, \textbf{Exogenous Drive} and \textbf{Attention-based Aggregation}. Specially, for an event of service $s_u$ at time $t$, updating $\mathbf{z}_{s_u}^t$ as:
\begin{equation}\label{eq:interacting_update}
    \mathbf{z}^{s_u}(t) = \sigma(\underbrace{\mathbf{W}_{a}\mathbf{h}^{\bar{t}}_{rec}(s_u)}_{\textbf{Aggregation}} + \underbrace{\mathbf{W}_{rec}\mathbf{z}^{\bar{t^{s_u}}}_{s_u}}_{Self-propagation} + \underbrace{\mathbf{W}_{t}(t-\bar{t^{s_u}})}_{\text{Exogenous Drive}})
\end{equation}
where $\mathbf{W}_{a} \in \mathbb{R}^{d_s \times d_s}$, $\mathbf{W}_{rec} \in \mathbb{R}^{d_s \times d_s}$ and $\mathbf{W}_{t} \in \mathbb{R}^{d_s}$ are learned parameters used to control the effect of above-mentioned three terms on the computation of service representation, respectively. $\sigma(\cdot))$ is a nonlinear function. $\mathbf{z}_{s_u}^{\bar{t^{s_u}}}$ is the previous representation of service $s_u$. $\bar{t}$ denotes the time point just before current event time $t$ and $\bar{t^{s_u}}$ represent the time point of last co-invocation event involved $s_u$. $\mathbf{h}^{\bar{t}}_{rec}(s_u) \in \mathbb{R}^d$ is the output representation obtained from the aggregation of service $s_u$'s neighbors $\mathcal{N}_{s_u}^{\bar{t}} = \{s_v : A_{s_u, s_v}^{\bar{t}} > 0\}$:
\begin{equation}
    \mathbf{h}^{\bar{t}}_{rec}(s_u) =  \max(\text{softmax}(\mathcal{S}_{s_u}^{\bar{t}})_{s_r}(\mathbf{W}_{h}\mathbf{z}_{s_r}^{\bar{t^u}}),  \forall s_r \in \mathcal{N}_{s_u}))
\end{equation}
where $\mathbf{W}_{h} \in \mathbb{R}^{d_s \times d_s}$ is learned parameters. Temporal attention $\mathcal{S}_{s_u}^{\bar{t}}$ is used to control the amount of information propagated from service $s_u$'s neighbors, which is updated by a hard-coded algorithm~\ref{alg:attention} adopted from DyRep\cite{trivedi2019dyrep}.

\begin{algorithm}[hbtp]
\caption{Update Algorithm for $\mathcal{S}$ and A}\label{alg:attention}
\SetKwInOut{Input}{Input}
\SetKwInOut{Output}{Output}

\Input{
    Co-invocation event record $o = (s_u, s_v, t)$ \; \\
    Co-invocation event intensity $\lambda_{s_u, s_v}^{t}$ \; \\
    Service representations of previous time $Z^{\bar{t}}$ \; \\
    Most recently updated $\mathbf{A}^{\bar{t}}$ and $\mathcal{S}^{\bar{t}}$ \; \\}

\Output{
    Updated  $\mathbf{A}^{t}$ and $\mathcal{S}^{t}$\;
}
$\mathcal{S}^{t} = \mathcal{S}^{\bar{t}}$ \\
$\mathbf{A}^{t} = \mathbf{A}^{\bar{t}}$\\
\uIf{$\mathbf{A}^{t}_{s_u, s_v} = 0$}{
         $\mathbf{A}^{t}_{s_u, s_v} = \mathbf{A}^{t}_{s_v, s_u} = 1$
    }
    
\For{\textbf{each} $j \in \{s_u,s_v\}$}{
    $b = \frac{1}{|\mathcal{N}_{j}^t|}$ \\
    $\mathbf{y} \gets \mathcal{S}^{t}_{j}$ \\
    \uIf{$\mathbf{A}^{\bar{t}}_{s_u, s_v} > 0$}{
        \tcc{$i$ is the other service involved in the event.} 
         $\mathbf{y}_i = b + \lambda_{s_u, s_v}^{t}$ \\
    }
    \Else{
        $b' = \frac{1}{|\mathcal{N}_{j}^{\bar{t}}|}$ \\
        $x = b' - b$ \\
        $\mathbf{y}_i = b + \lambda_{s_u, s_v}^{t}$ \\
        $\mathbf{y}_w = \mathbf{y}_w - x \quad \forall w \ne i, \mathbf{y}_w \ne 0$ \\
    }
    Normalize $\mathbf{y}$ and set $ \mathcal{S}^{t}_{j} \gets \mathbf{y}$ \\
}

\Return $\mathbf{A}^{t}$ and $\mathcal{S}^{t}$
\end{algorithm}

Conditional intensity $\lambda_{s_u, s_v}^{t}$ modes the occurrence of co-invocation between $s_u$ and $s_v$ at time $t$:
\begin{equation}
    \lambda_{s_u, s_v}^{t} = \psi\log(1 + \exp\{\frac{\omega^{\intercal}[\mathbf{z}_{s_u}^{\bar{t}};\mathbf{z}_{s_v}^{\bar{t}}]}{\psi}\})
\end{equation}
where $\psi$ is trainable scalar parameter, which denotes the rate of events arising from a point process, and $\mathbf{\omega} \in \mathbb{R}^{2d_s}$ is designed to learn time-scale specific compatibility. $[;]$ denotes concatenation.

\subsubsection{Loss Function}
For a given set $\mathcal{O}$ of $\mathcal{P}$ co-invocation events, we learn parameters $\Omega=\{\mathbf{W}_{a}, \mathbf{W}_{rec}, \mathbf{W}_{t}, \mathbf{W}_{h}, \psi, \omega \}$ by minimizing the following loss function:
\begin{equation}\label{eq:loss2}
    \mathcal{L}_{2} = \mathcal{L}_{events} + \mathcal{L}_{nonevents}
\end{equation}
where $\mathcal{L}_{events} = -\sum_{p}^{\mathcal{P}}\log(\lambda_{o_p})$ is the total negative log of the intensity rate for all co-invocation events between service $s_u^p$ and service $s_v^p$; $\mathcal{L}_{nonevents} = \int_{0}^{T}\Gamma(\tau)d\tau = \sum_{m}^{\mathcal{M}}\log(\lambda_{o_m})$ represent total survival probability for co-invocation events that do not happen. It is intractable to compute all non-positive nonevent, we use Monte Carlo method to sample a subset to compute this term, with following \cite{trivedi2019dyrep} setting $\mathcal{M}=5\mathcal{P}$.

\subsection{GAN-style training step}\label{sec:gan-style}
To model the process that requirements trigger service evolution and that service evolution reacting the selection of services by requirements, we adopt a GAN-style training step. Specifically, we use two independent Adam optimizers\cite{kingma2014adam} $opt_1$ and $opt_2$ to optimize the parameters $\{\Psi, \Lambda\}$ in the supervised service recommendation task and parameters $\Omega$ in the unsupervised evolving service representation task.  It is also important to note that we perform supervised task optimization for each batch of samples before unsupervised task optimization. . The pseudo-code for the GAN-style training algorithm is given in Algorithm~\ref{alg:training}.

\begin{algorithm}[hbtp]
\caption{Traning Algorithm for DySR}\label{alg:training}
\SetKwInOut{Input}{Input}
\SetKwInOut{Output}{Output}

\Input{
    Requirement sample set $Y$ \; \\
    Number of epochs $n$ \; \\
    Parameter sets $\{\Psi, \Lambda, \Omega\}$ \; \\
    Optimizers $opt_1$ and $opt_2$ \; \\
}

\Output{
    Updated parameter sets $\{\Psi, \Lambda, \Omega\}$\;
}
\For{epoch = $1, \dots , n$}{
    \For{\textbf{each} requirement $(\mathbf{v}_r, C+) \in Y$}{
        \tcc{Optimize supervised service recommendation task related parameters}
        Do forward step described in Section~\ref{sec:supervised} \\
        Update $\{\Psi, \Lambda\}$ to minimize $\mathcal{L}_1$ in Eq.~\ref{eq:loss1} with $opt_1$ \\
        
        \tcc{Optimize unsupervised unsupervised evolving service dynamic task related parameters}
        Generate co-invocation event set $\mathcal{O}$ from $C^+$ \\
        \For{\textbf{each} co-invocation evnet $(s_u, s_v, t) \in \mathcal{O}$}{
            Do forward step described in Section~\ref{sec:unsupervised} \\
            Update $\Omega$ to minimize $\mathcal{L}_2$ in Eq.~\ref{eq:loss2} with $opt_2$ \\
        }
    }
}

\Return $\{\Psi, \Lambda, \Omega\}$
\end{algorithm}

\section{Experiment Settings}\label{sec:settings}
\subsection{Dataset \& Metrics}
We evaluate the proposed DySR model on the real-world \textbf{ProgrammableWeb} dataset, which is also the dataset used in existing service recommendation studies. 

\textbf{ProgrammableWeb}: The dataset is the largest online Web service registry. We collected a total of 23,520 APIs and 7,947 mashups on Oct 10, 2020. The mashups without functional description, the services that have not been invoked, and the mashups with fewer than two component services were removed. The experimental dataset contains 3,380 mashups, whose functional descriptions are used as requirements, and 720 APIs. We sorted the mashups by when they were created, and we initialize the adjacency matrix $A$ using the co-invocations from the earliest $300$ mashups. The next $2,400$ mashups are used as the training set, and the remaining $680$ mashups are used as the test set.

We adopted the following evaluation metrics to measure the recommendation performance:
\begin{equation}
    Precision@N = \frac{1}{|R|}\sum_{r \in R}\frac{|\hat{C_r}\cap C_r|}{|\hat{C_r}|}
\end{equation}

\begin{equation}
    Recall@N = \frac{1}{|R|}\sum_{r \in R}\frac{|\hat{C_r}\cap C_r|}{|C_r|}
\end{equation}

\begin{equation}
    F1@N = = \frac{1}{|R|}\sum_{r \in R}\frac{|\hat{C_r}\cap C_r|}{|C_r| + |\hat{C_r}|}
\end{equation}
where $R$ is the set of requirements in the test set and $|R|$ denotes the size of $R$. For requirement $r$, $\hat{C_r}$ is the recommended services, while $C_r$ is its actual component services. 

\subsection{Implementation Details}
We use \textit{bert-base-uncased} provided by Transformers\cite{wolf-etal-2020-transformers} to obtain requirement representation with the dimension $d_r$ set to $768$, and it should be noted that we do not do fine-tune on \textit{bert-base-uncased}. Service representations $Z^0$ are initialized by a $128d$ Word2Vec\cite{mikolov2013word2vec} word embedding trained on \textit{text8}\footnote{http://mattmahoney.net/dc/text8.zip}. We use the two different pre-trained language models to reflect the representation gap between requirement and service. It should be noted that the initial representation of the service in the DySR model can theoretically be obtained  using other information or in a random way. We implement a DySR variant called DySR-Rand, which does not need any prior knowledge of service and uses a randomly initialized representation of the service.

Gradient clipping is used in $opt_2$ to avoid gradient explosion, and the clipping value is set to $100$. We do not use dropout, and batch size is set to $50$. We conduct five independent experiments for each  approach to preventing serendipity, and early-stop is applied to avoid over-fitting. All the results reported are average results.

\begin{figure*}[!htbp]
    \centering
    \subfigure[Precision@N]{
    \includegraphics[width=0.45\linewidth]{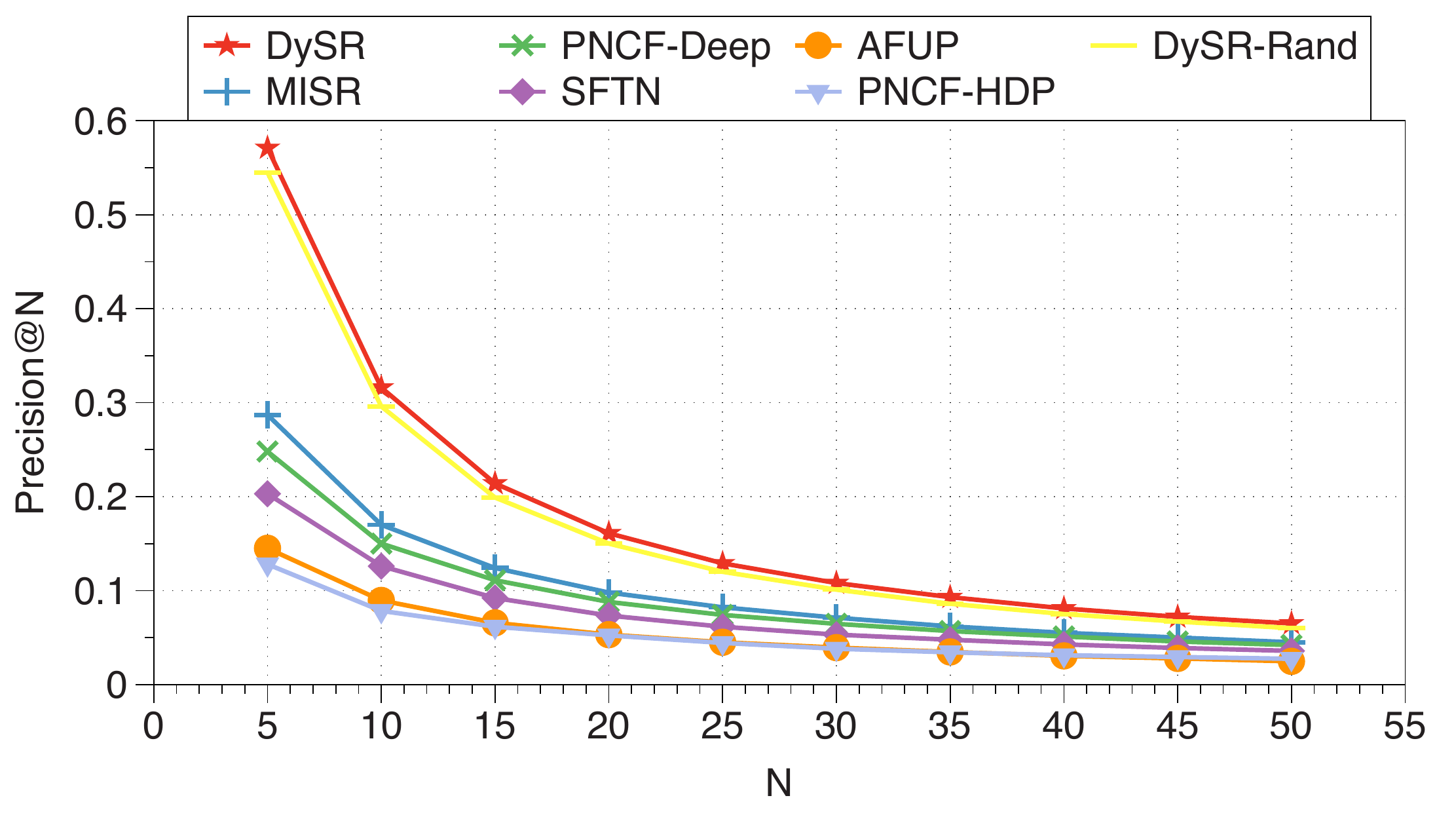}
    }
    \subfigure[Recall@N]{
    \includegraphics[width=0.45\linewidth]{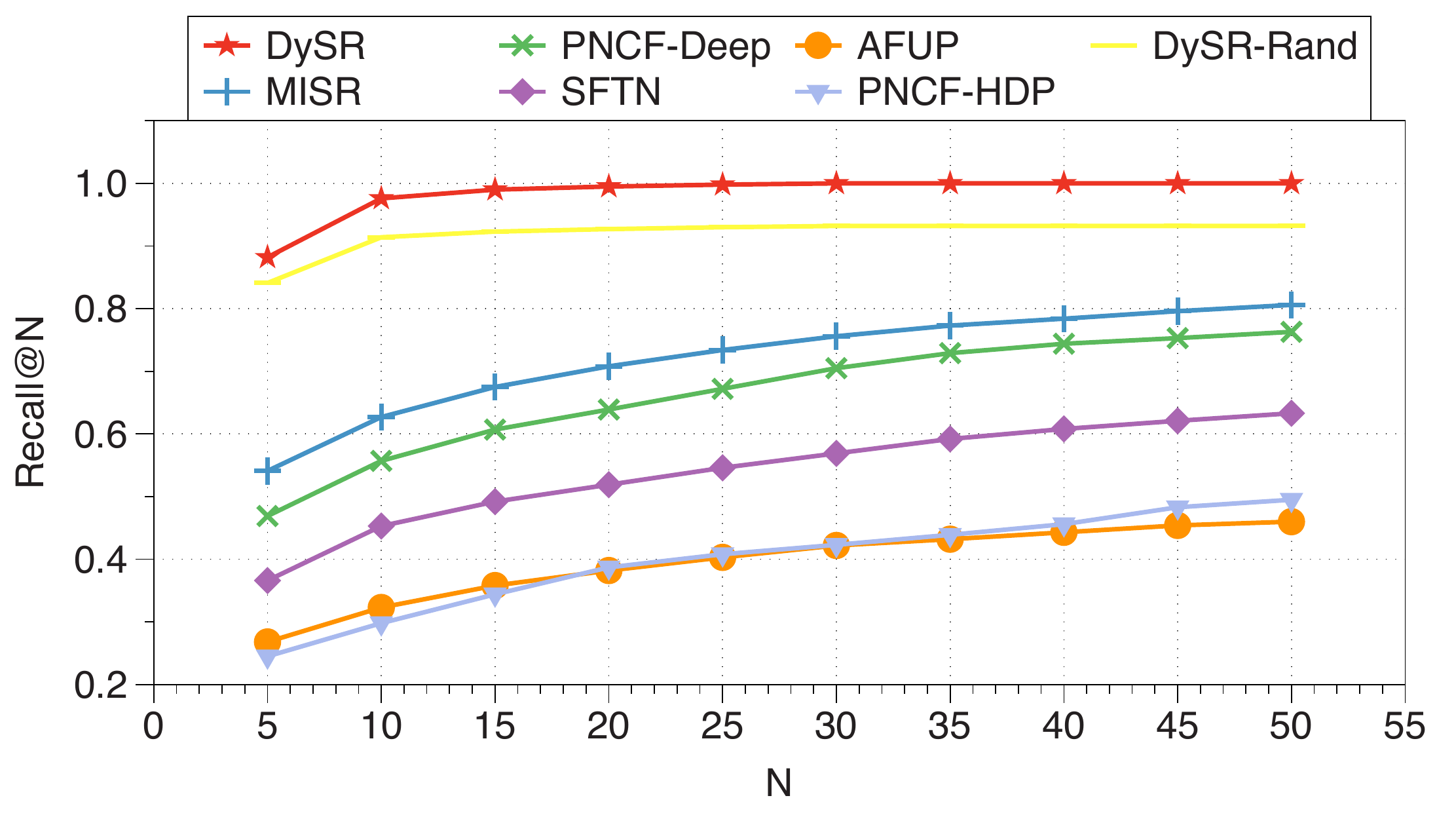}
    }
    \subfigure[F1@N]{
    \includegraphics[width=0.45\linewidth]{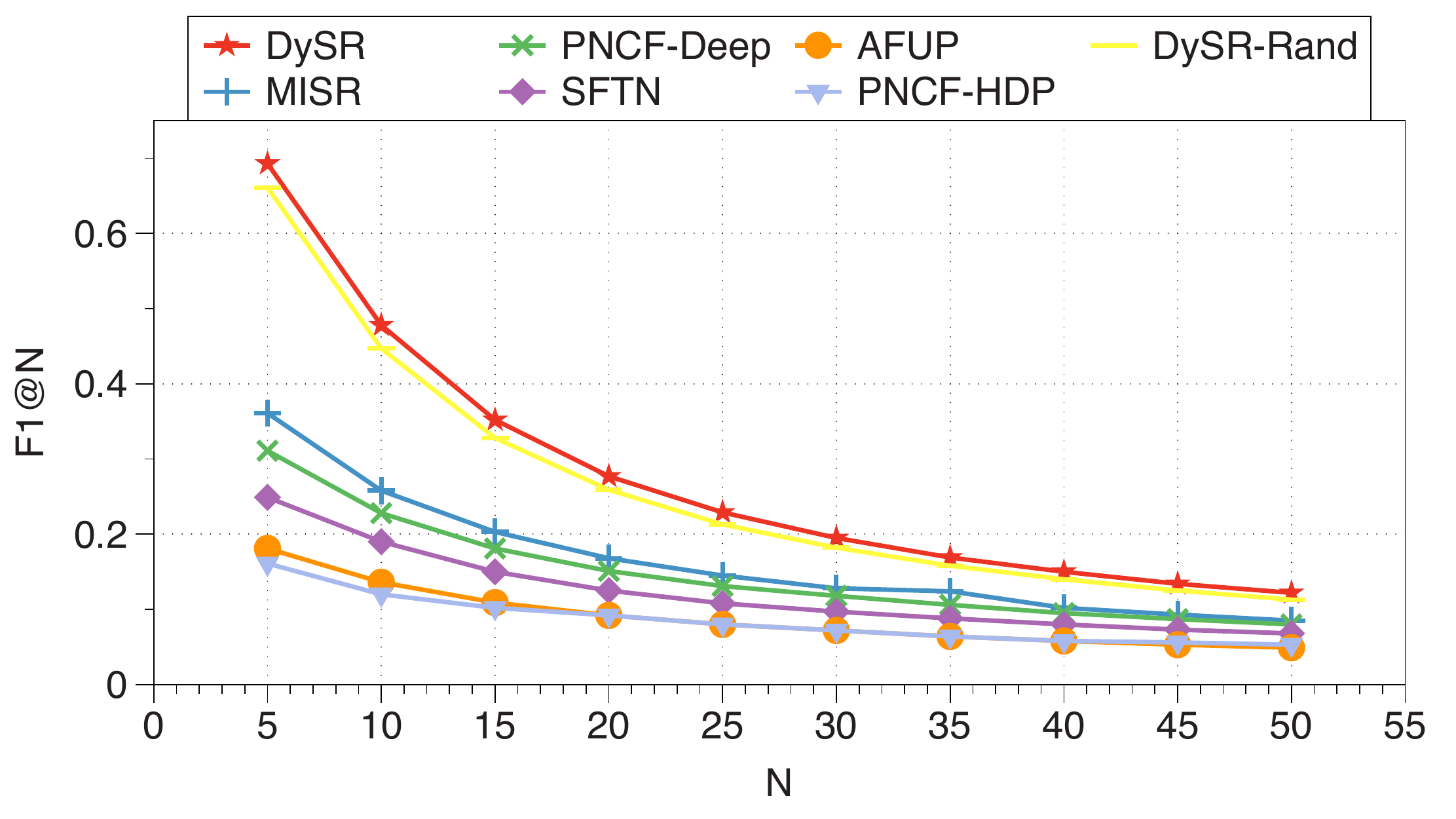}
    }
    
    \caption{Performance comparison of different approaches.}
    \label{fig:baseline}
\end{figure*}

\subsection{Baselines}
To evaluate the effectiveness of model, we select five state-of-the-art service recommendation approaches:
\begin{enumerate}
    \item AFUP\cite{jain2015aggregating}: This approach first leverage probabilistic topic models to compute relevance between a service and a given requirement. description. And then use collaborative filtering to estimate the probability of a service being used by existing similar requirements. Finally, the multiplies these two term based on Baye's theorem to rank candidates service.
    \item SFTN\cite{samanta2017recommending}: This approach extend AFUP by using hierarchical dirichlet process (HDP) and probabilistic matrix factorization (PMF) to tackle cold start issues and usage history.
    \item PNCF\cite{chen2018software}: This approach use multi-layer perceptron to capture the non-linear user-item relationships and obtain abstract data representation from sparse vectors. However, text features were not considered in their original version, so we constructed two variants: PNCF-HDP using HDP adopted in SFTN to obtain text features, and PNCF-Deep using a pre-trained language model to obtain text features.
    \item MISR\cite{ma2021deep}: This approach propose a deep neural network that can captures multiplex interactions between services and requirements to extract hidden structures and features for better recommendation performance.
\end{enumerate}

It should be noted that most baselines do not provide official code, and we can only reproduce it as described in their papers. In some cases, we were not able to reproduce the results they reported, possibly due to different ways of dividing the dataset and missing important parameter values. In these cases, we chose to directly compare the results reported in \cite{ma2021deep}.

\section{Results \& Discussion}\label{sec:results}
\subsection{Overview}
In this section, we give an overview performance comparison between our proposed DySR and baseline approaches. Fig.~\ref{fig:baseline} shows the performance comparison of different approaches, showing that the DySR and DySR-Rand outperform all the five baselines across all evaluation metrics.

The AFUP and PNCF-HDP performed the worst of all the baselines for the following two main reasons: 1) They use a topic model to extract service/requirement representation from the description text, which ignores the order of words and further leads to lost semantic information; 2) Rough handling of service historical usage information. The introduction of probabilistic matrix factorization allows SFTN to handle historical usage information somewhat better, but the poor service/requirement representations obtained by the topic model remain limiting to its performance. The PNCF-Deep and MISR perform better than the other baselines because they use a pre-trained language model to obtain better representations of the services/requirements. In addition, MISR performs the best of all the benchmark methods because it takes into account multiple types of interactions between services and requirements.

Although DySR-Rand does not require any a priori knowledge of the service. DySR-Rand obtains competitive results with DySR, which shows that our proposed model is well suited to eliminate the gap between requirements and services.

DySR and DySR-Rand require less information than the baselines, while their performance is significantly better than all baselines. For example, all baselines need to maintain a requirement-service invocation matrix, while DySR only needs to maintain a small-scale service co-invocation matrix, and MISR additionally requires the tags of the service and historical requirements, which are not needed in DySR. Compared to the best performing 
baseline approach MISR, which requires the most information, DySR (DySR-Rand) improves the $Precision@5$, $Recall@5$ and $F1@5$ metrics by \textbf{28.4\% (25.8\%)}, \textbf{34.1\% (30\%)} and \textbf{33.2\% (30\%)}, respectively. The performance improvements mainly benefit from evolving service representation and transformation function that allows the use of time information and the acquisition of better representation of services and requirements in same vector space. 

\begin{figure*}[!htbp]
    \centering
    \subfigure[Precision@N]{
    \includegraphics[width=0.45\linewidth]{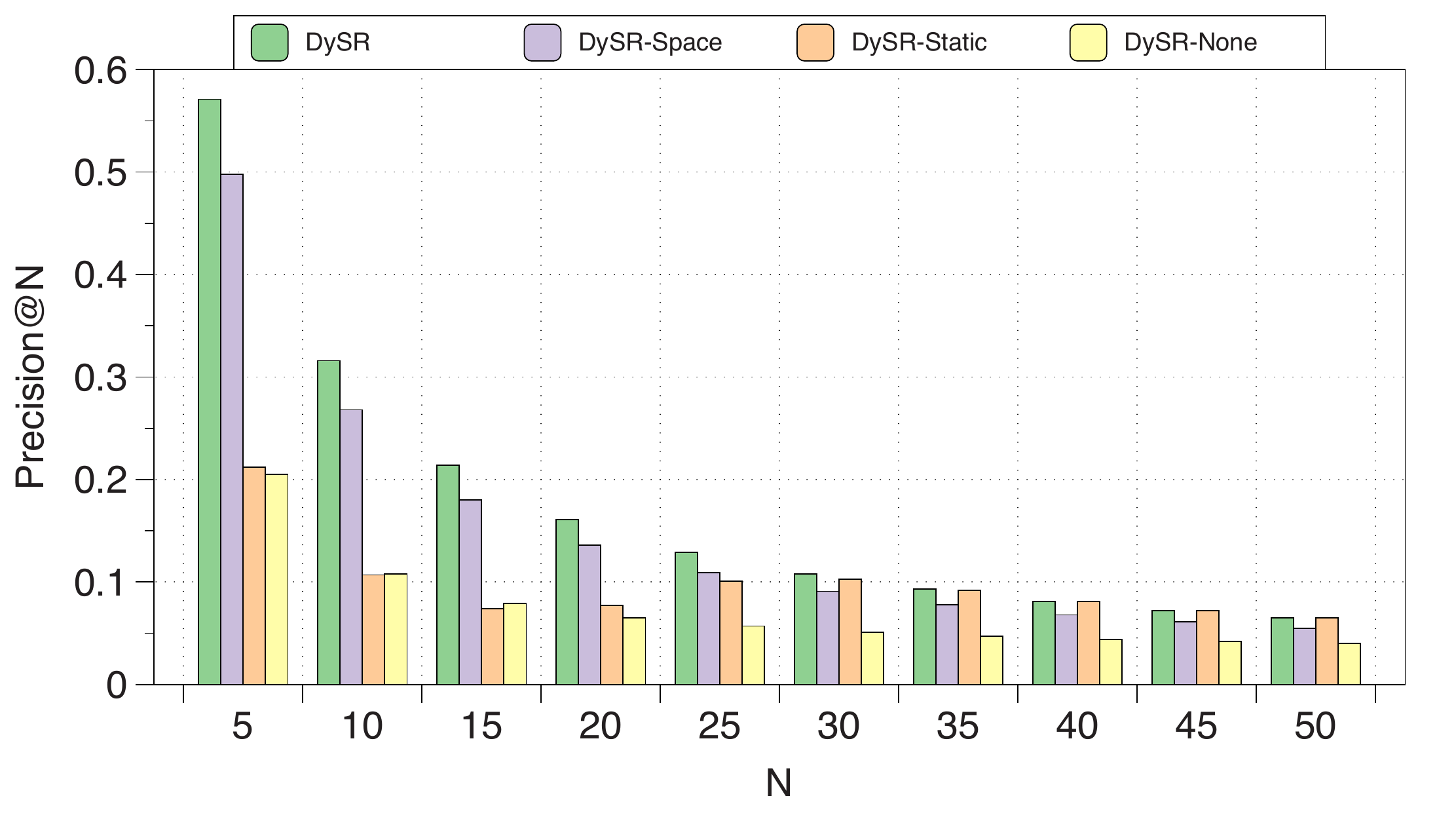}
    }
    \subfigure[Recall@N]{
    \includegraphics[width=0.45\linewidth]{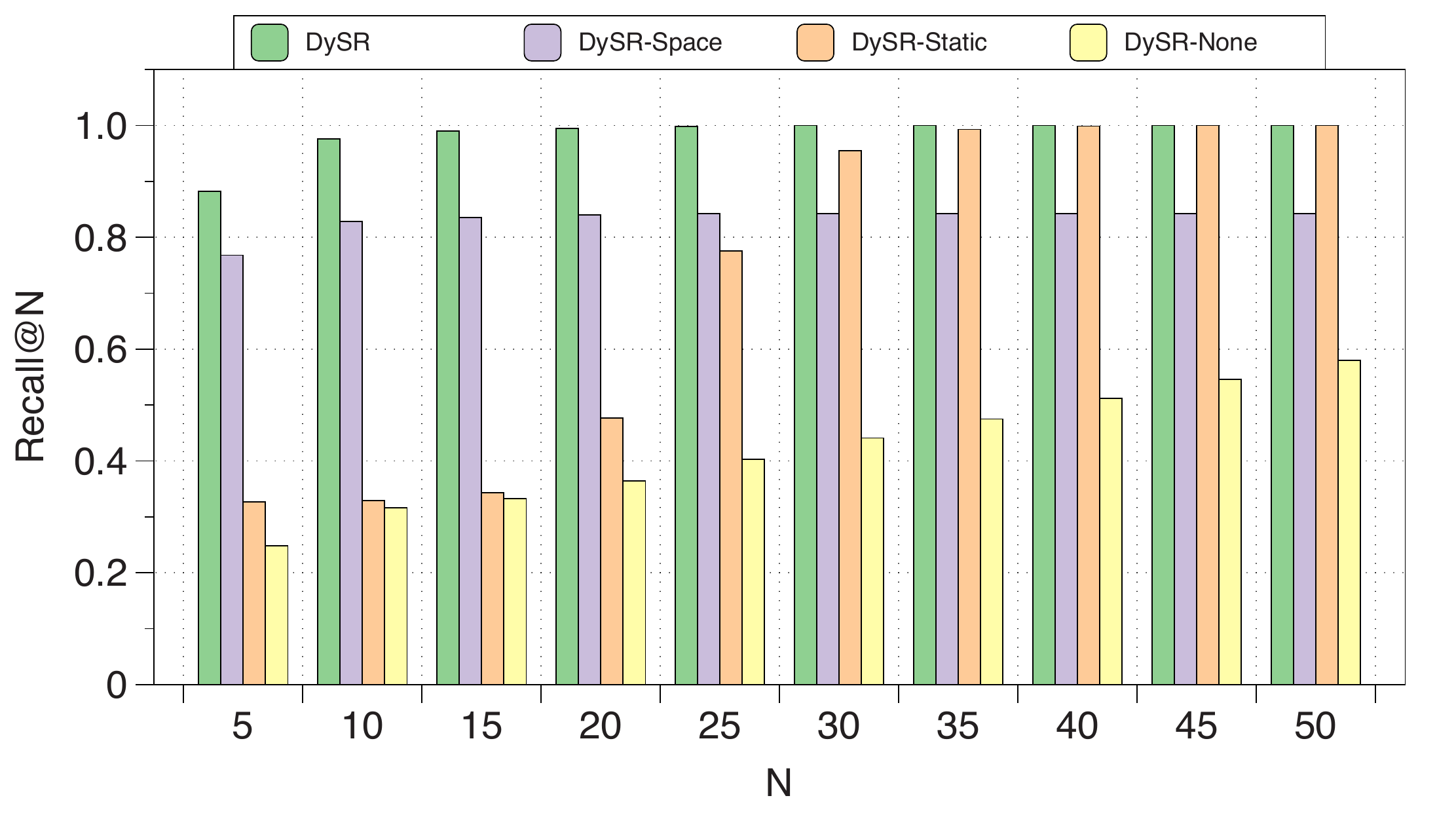}
    }
    \subfigure[F1@N]{
    \includegraphics[width=0.45\linewidth]{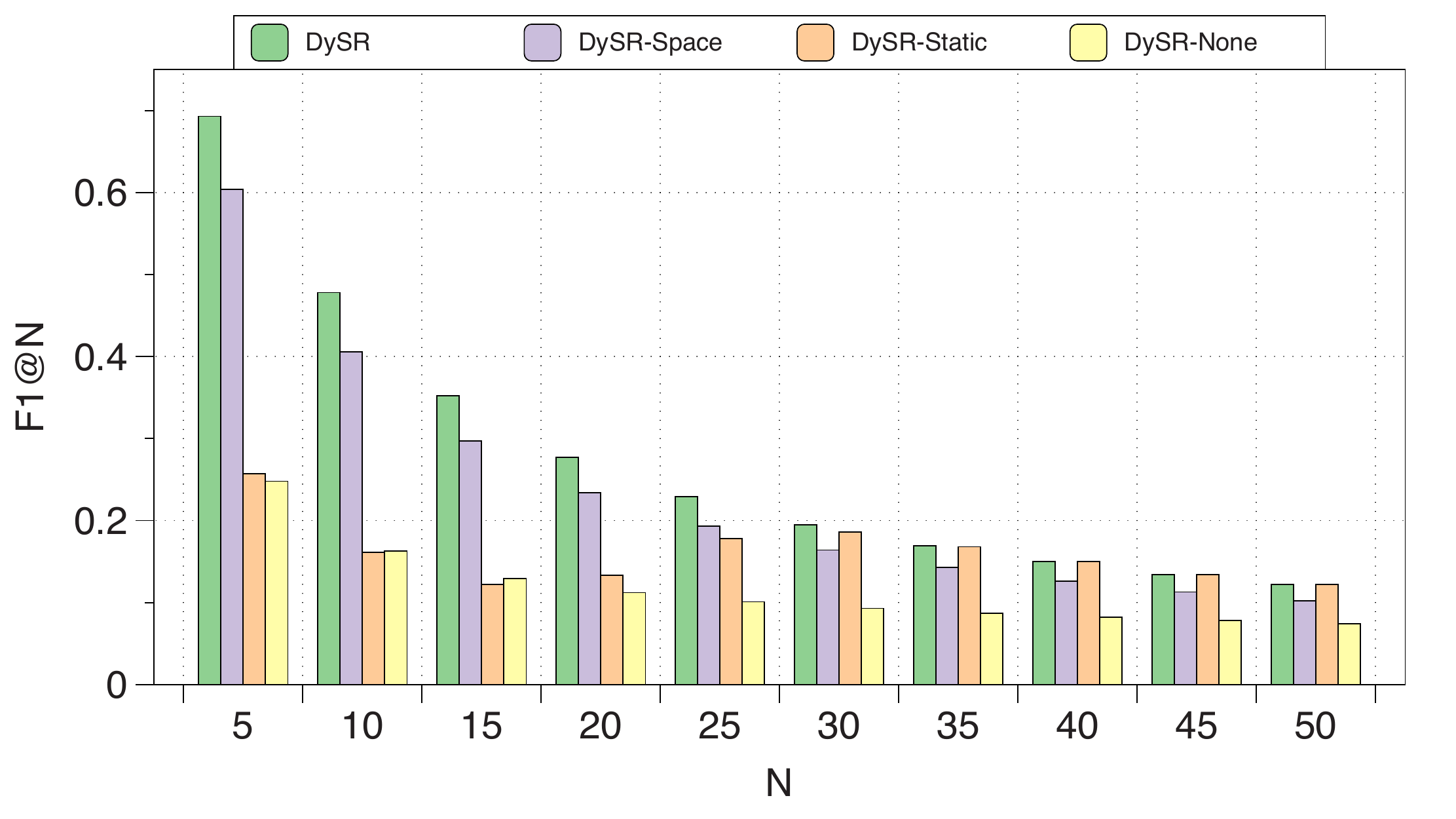}
    }
    
    \caption{Performance comparison of different variants of DySR.}
    \label{fig:variant}
\end{figure*}
\subsection{Ablation Study \& Qualitative Performance}

DySR unifies several components that contribute to its effectiveness in service recommendation. In this section, we provide insights on \textbf{evolving service representation} component and \textbf{transformation} component and how they are indispensable to the service recommendation by performing an ablation study on various design choices of DySR. We designed the following variants of DySR for comparison:
\begin{itemize}
    \item \textbf{DySR-Static}: In this variant, we do not perform unsupervised evolving service representation tasks so that the service representation remains initialised, i.e. $Z^{t} \equiv Z^0$.
    \item \textbf{DySR-Space}: In this variant, we turn off transformation function (Eq.~\ref{eq:transformation})  and directly use the original requirement representation and the dynamic service representation as inputs to the requirement-service matching function (Eq.~\ref{eq:rs}). The dimension of the corresponding requirement-service matching matrix  $W_{\Lambda}$ sets to $d_r \times d_s$.
    \item \textbf{DySR-None}: In this variant, we do not perform unsupervised evolving service representation task and turn off transformation function.
\end{itemize}
The comparison among the variants of our approach is shown in Fig.~\ref{fig:variant}. Both DySR-Space and DySR-Static outperform DySR-None on $Precision@N$, $Recall@N$ and $F1@N$,  suggesting that \textbf{evolving service representation} and \textbf{transformation function} do contribute to the recommended performance. DySR outperforms DySR-Space and DySR-static indicating that \textbf{evolving service representation} and \textbf{transformation function} are complementary. 

\begin{figure*}[!htbp]
    \centering
    \subfigure[Static Service Representation]{
    \includegraphics[width=0.45\linewidth]{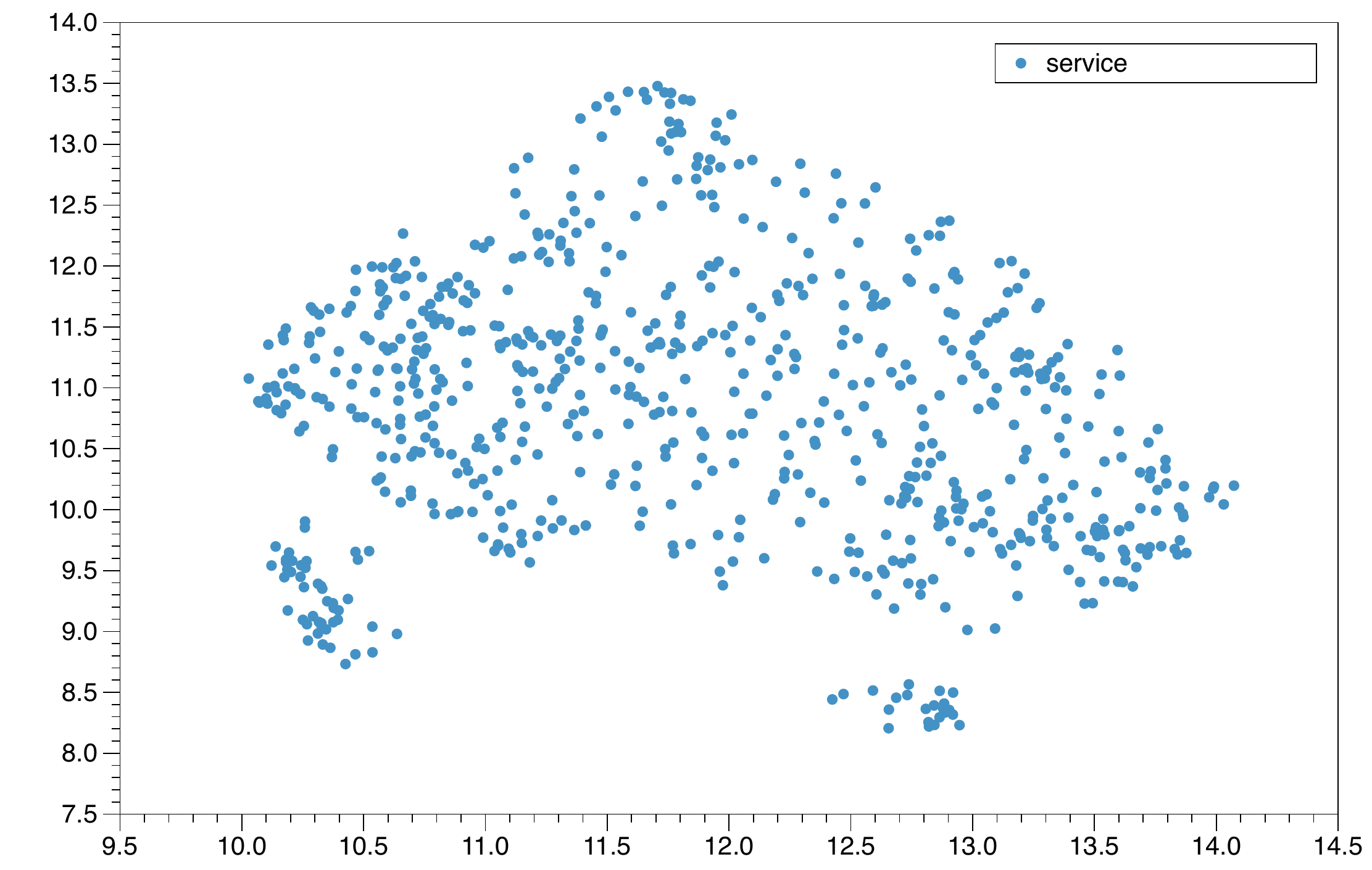}
    }
    \subfigure[Evolving Service Representation]{
    \includegraphics[width=0.45\linewidth]{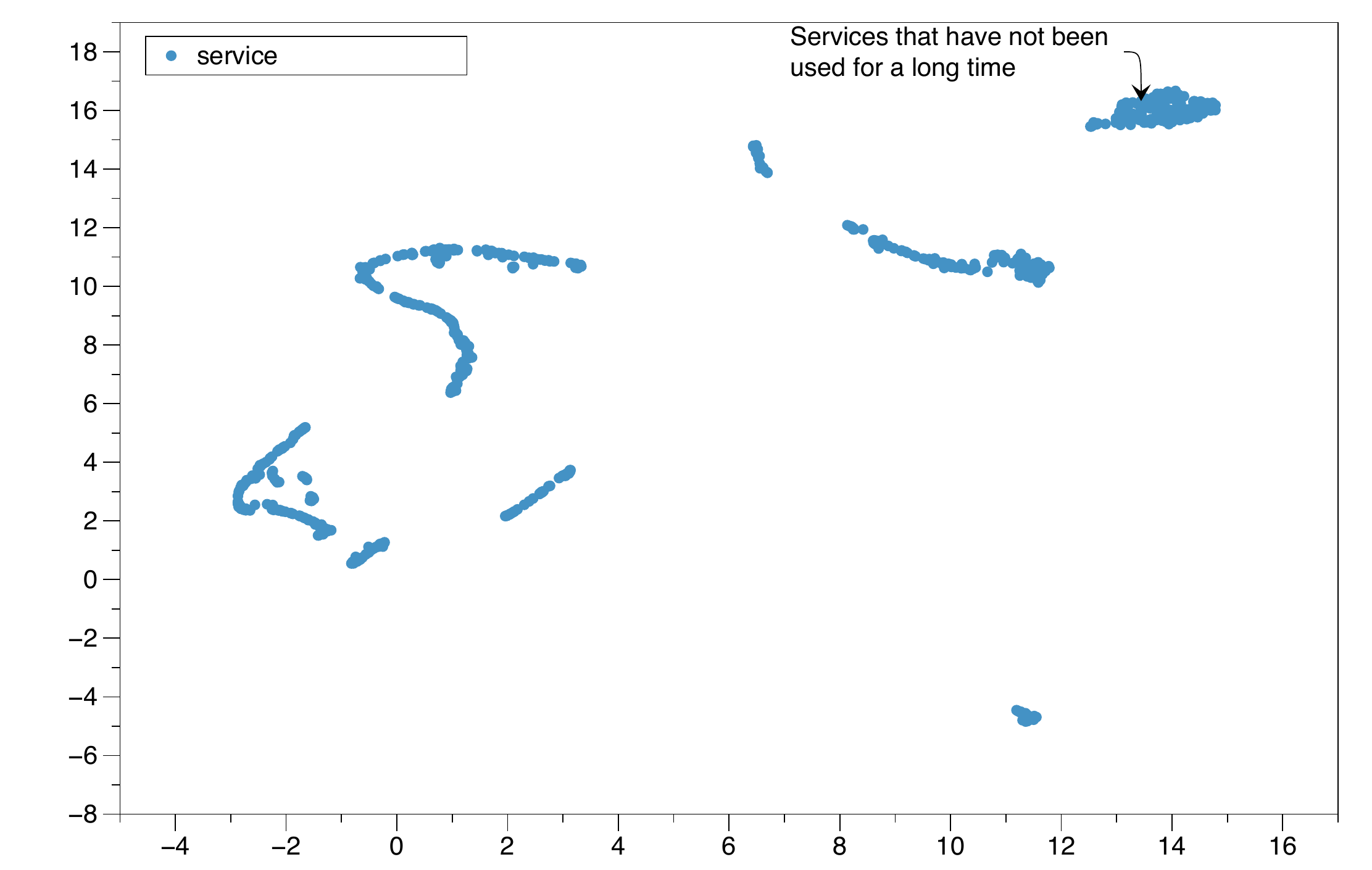}\label{fig:s_b}
    }
    
    \caption{UMAP for static service representation and evolving service representation.}
    \label{fig:service_space}
\end{figure*}
We conduct a series of qualitative analyses to understand how evolving service representation and transformation function contribute to recommendation performances. We first compare our evolving service representation against the static content-based service representation. Fig.~\ref{fig:service_space} shows the UMAP\cite{mcinnes2018umap-software} embeddings leaned by DySR (right) and initialised by service function description. The visualization demonstrates the evolving service representation have more discriminative power as it can effective capture the distinctive and evolving patterns of service combinations (line clusters in Fig.~\ref{fig:s_b}) as well as outdated sets of services (block clusters in Fig.~\ref{fig:s_b}) with empirical evidence.

\begin{figure*}[!htbp]
    \centering
    \subfigure[Static Services \& Origin Requirements (DySR-None)]{
    \includegraphics[width=0.45\linewidth]{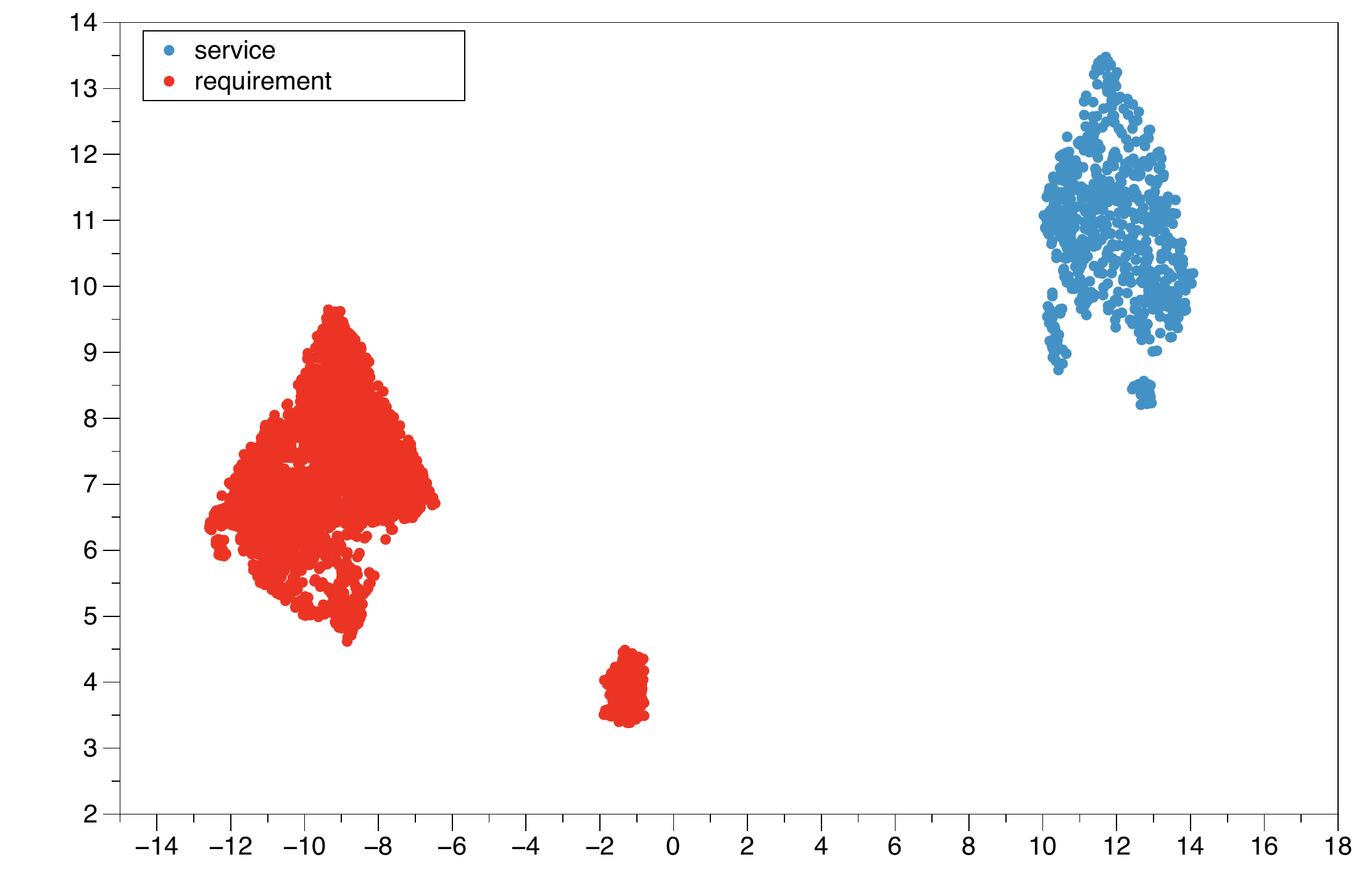}\label{fig:space_a}
    }
    \subfigure[Static Services \& Aligned Requirements (DySR-Static)]{
    \includegraphics[width=0.45\linewidth]{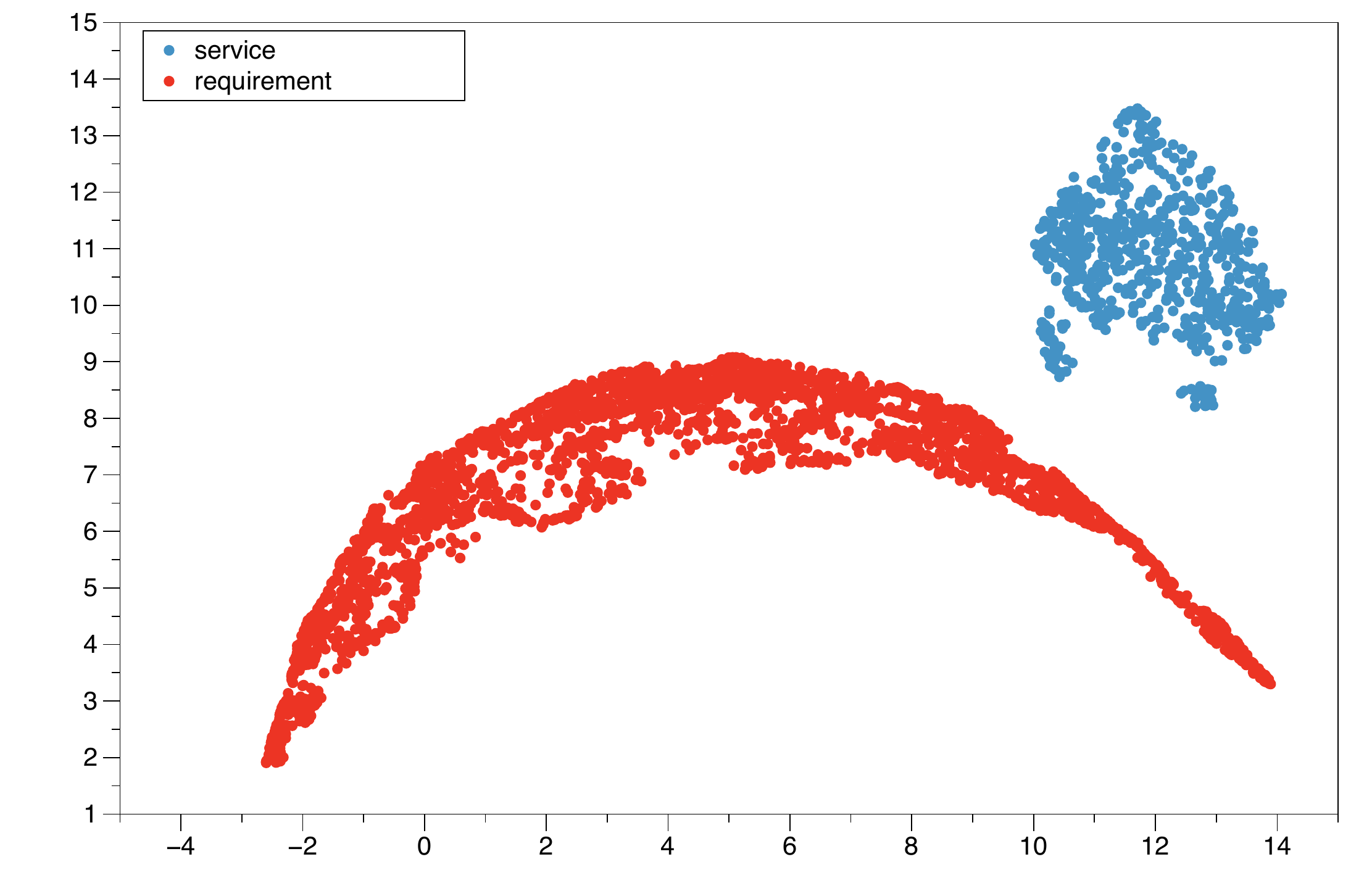}\label{fig:space_b}
    }
    \subfigure[Evolving Services \& Origin Requirements (DySR-Space)]{
    \includegraphics[width=0.45\linewidth]{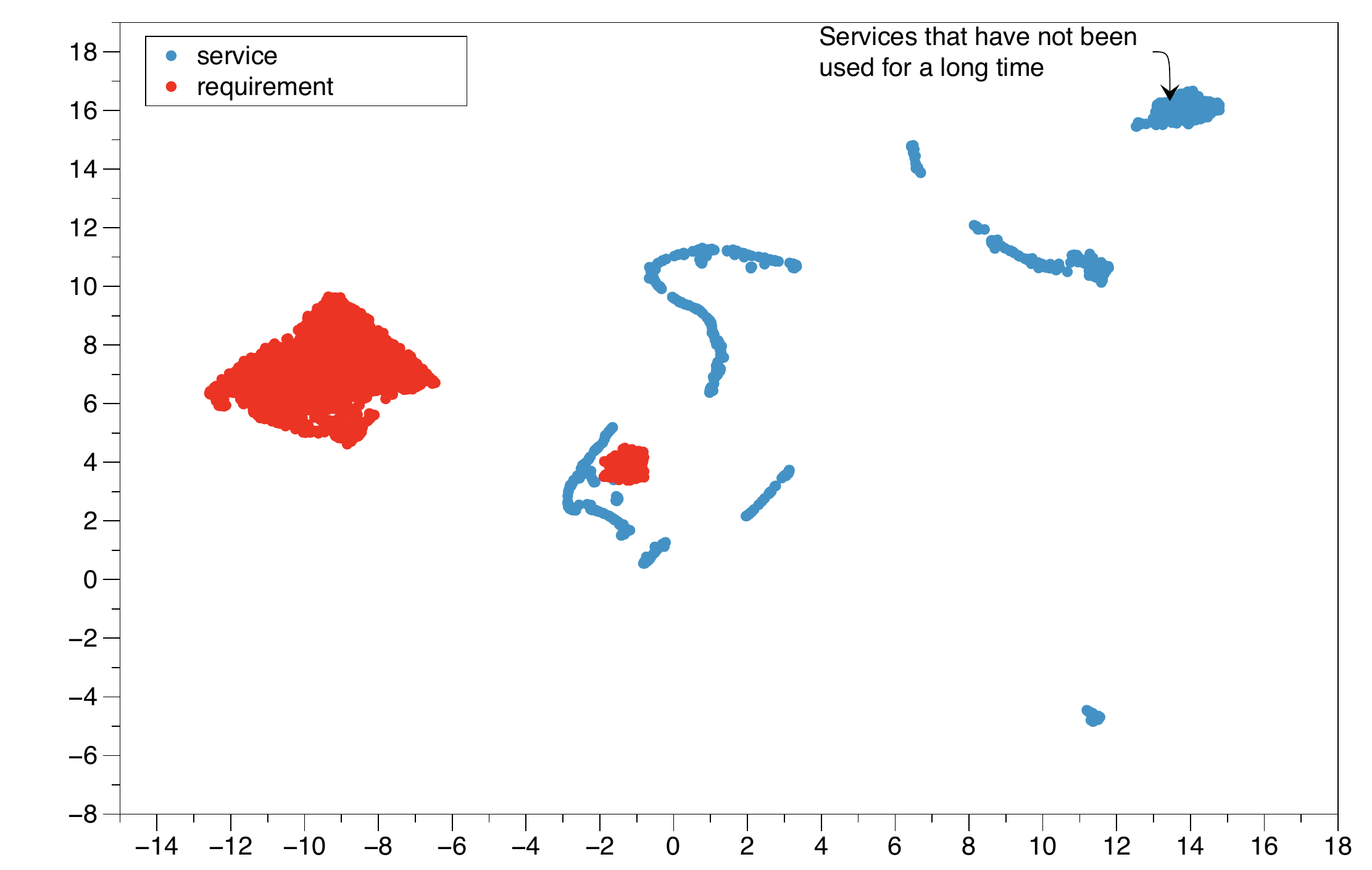}\label{fig:space_c}
    }
    \subfigure[Evolving Services \& Aligned Requirements (DySR)]{
    \includegraphics[width=0.45\linewidth]{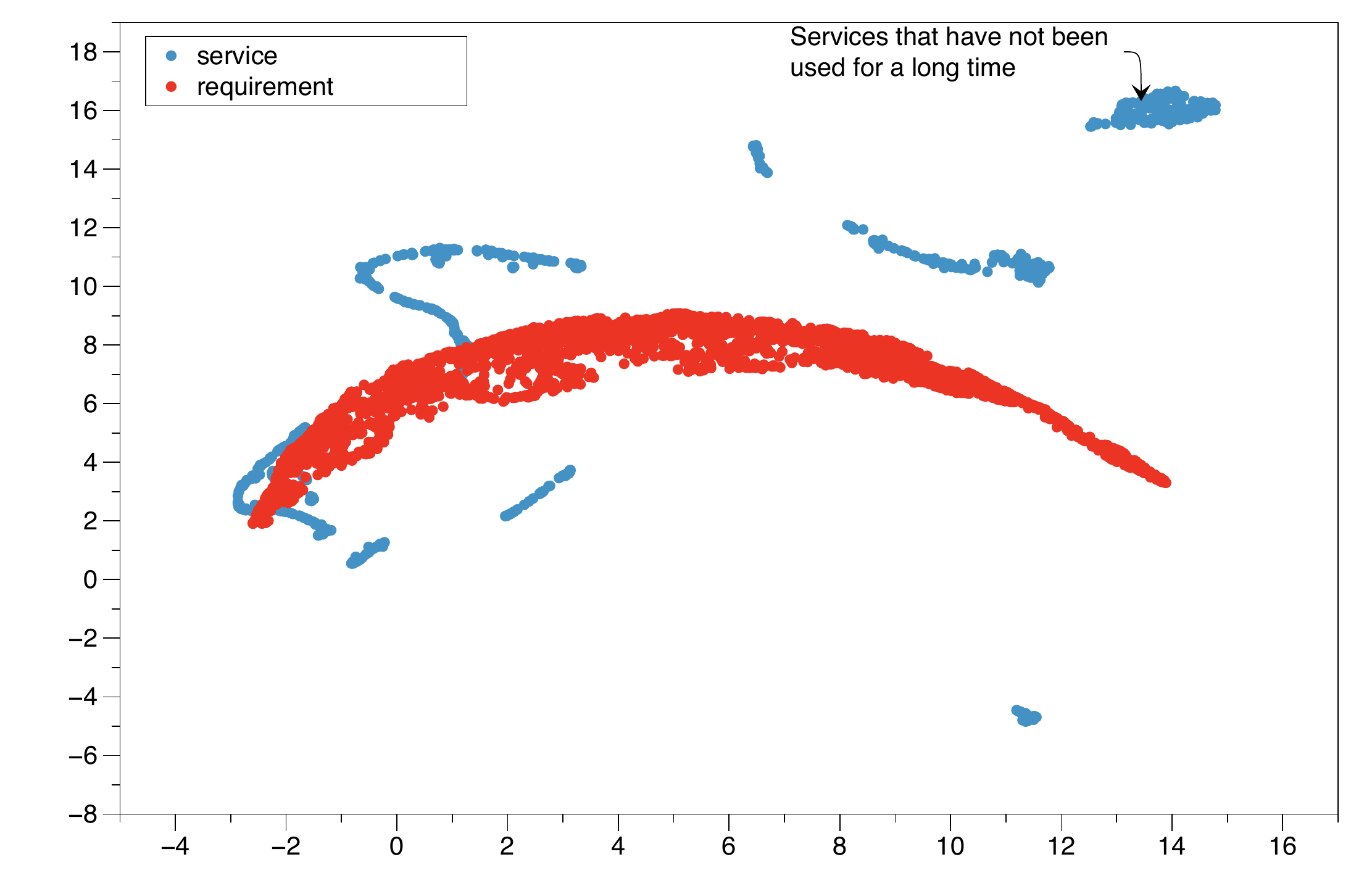}\label{fig:space_d}
    }
    \caption{UMAP for explaining how DySR eliminates the representation gap between service and requirement.}
    \label{fig:space}
\end{figure*}
Fig.~\ref{fig:space} explains how DySR eliminates the representation gap between service and requirement. Fig.~\ref{fig:space_a} demonstrates the gap between the representation of requirements and the representation of service that we stated in Section~\ref{sec:intro}. Compared with Fig.~\ref{fig:space_a}, the requirement representation space is much closer to the space of service representation in Fig~\ref{fig:space_b}, which shows that the transformation function in DySR does eliminate the difference between requirement and service representations to some extent. The same conclusion can be obtained by comparing Fig.~\ref{fig:space_c} and Fig.~\ref{fig:space_d}. Fig.~\ref{fig:space_c} and Fig.~\ref{fig:space_d} can also illustrate that evolving service representations reduce the probability of selecting outdated services by moving them away from requirement representation space thereby improving recommendation performance.


%

%

    
\section{Conclusion}
In this paper, we propose an end-to-end deep learning model called DySR for cold-start service recommendations. DySR solves the service evolution problem by introducing evolving service representation, eliminating the gap between services and requirements through transformation functions, and solving the cold start problem through a requirement-service matching function. Experiments on a real-world dataset demonstrated that the proposed approach significantly outperforms several state-of-the-art service recommendation methods regarding three evaluation metrics. In 
future work, we will try to show theoretically why evolving service representation and space alignment have an impact on service performance.

\section*{Acknowledgment}
The research in this paper is partially supported by the National Key Research and Development Program of China (No 2018YFB1402500) and the National Science Foundation of China (61772155, 61832004, 61802089, 61832014).

\ifCLASSOPTIONcaptionsoff
  \newpage
\fi



\bibliographystyle{IEEEtran}
\bibliography{reference.bib}

\begin{thebibliography}{10}
\providecommand{\url}[1]{#1}
\csname url@samestyle\endcsname
\providecommand{\newblock}{\relax}
\providecommand{\bibinfo}[2]{#2}
\providecommand{\BIBentrySTDinterwordspacing}{\spaceskip=0pt\relax}
\providecommand{\BIBentryALTinterwordstretchfactor}{4}
\providecommand{\BIBentryALTinterwordspacing}{\spaceskip=\fontdimen2\font plus
\BIBentryALTinterwordstretchfactor\fontdimen3\font minus
  \fontdimen4\font\relax}
\providecommand{\BIBforeignlanguage}[2]{{%
\expandafter\ifx\csname l@#1\endcsname\relax
\typeout{** WARNING: IEEEtran.bst: No hyphenation pattern has been}%
\typeout{** loaded for the language `#1'. Using the pattern for}%
\typeout{** the default language instead.}%
\else
\language=\csname l@#1\endcsname
\fi
#2}}
\providecommand{\BIBdecl}{\relax}
\BIBdecl

\bibitem{WU2015133}
W.~Zhaohui, D.~Shuiguang, and W.~Jian, ``Chapter 6 - service recommendation
  service computing,'' W.~Zhaohui, D.~Shuiguang, and W.~Jian, Eds.\hskip 1em
  plus 0.5em minus 0.4em\relax Boston: Academic Press, 2015, pp. 133--176.

\bibitem{faieq2019context}
S.~Faieq, A.~Front, R.~Saidi, H.~El~Ghazi, and M.~D. Rahmani, ``A context-aware
  recommendation-based system for service composition in smart environments,''
  \emph{Service Oriented Computing and Applications}, vol.~13, no.~4, p.
  341–355, 2019.

\bibitem{lei2015time}
Y.~Lei, Z.~Jiantao, Z.~Junxing, W.~Fengqi, and W.~Juan, ``Time-aware semantic
  web service recommendation,'' in \emph{2015 IEEE International Conference on
  Services Computing}, 2015, pp. 664--671.

\bibitem{lin2018nl2api}
C.~Lin, A.~Kalia, J.~Xiao, M.~Vukovic, and N.~Anerousis, ``Nl2api: A framework
  for bootstrapping service recommendation using natural language queries,'' in
  \emph{2018 IEEE International Conference on Web Services (ICWS)}, 2018, pp.
  235--242.

\bibitem{hu2019poi}
S.~Hu, Z.~Tu, Z.~Wang, and X.~Xu, ``A poi-sensitive knowledge graph based
  service recommendation method,'' in \emph{2019 IEEE International Conference
  on Services Computing (SCC)}, 2019, pp. 197--201.

\bibitem{chang2021graph}
Z.~Chang, D.~Ding, and Y.~Xia, ``A graph-based qos prediction approach for web
  service recommendation,'' \emph{Applied Intelligence}, pp. 1--15, 2021.

\bibitem{cao2019efficient}
Z.~Cao, X.~Qiao, S.~Jiang, and X.~Zhang, ``An efficient knowledge-graph-based
  web service recommendation algorithm,'' \emph{Symmetry}, vol.~11, no.~3, p.
  392, 2019.

\bibitem{gao2016seco}
Z.~Gao, Y.~Fan, C.~Wu, W.~Tan, J.~Zhang, Y.~Ni, B.~Bai, and S.~Chen,
  ``Seco-lda: Mining service co-occurrence topics for recommendation,'' in
  \emph{2016 IEEE International Conference on Web Services (ICWS)}, 2016, pp.
  25--32.

\bibitem{ma2021deep}
Y.~Ma, X.~Geng, and J.~Wang, ``A deep neural network with multiplex
  interactions for cold-start service recommendation,'' \emph{IEEE Transactions
  on Engineering Management}, vol.~68, no.~1, p. 105–119, 2021.

\bibitem{gu2021csbr}
Q.~Gu, J.~Cao, and Y.~Liu, ``Csbr: A compositional semantics-based service
  bundle recommendation approach for mashup development,'' \emph{IEEE
  Transactions on Services Computing}, 2021.

\bibitem{wang2021external}
Z.~Wang, M.~Liu, Z.~Tu, and X.~Xu, ``External service sensing (ess): Research
  framework, challenges and opportunities,'' 2021.

\bibitem{bai2017sr}
B.~Bai, Y.~Fan, W.~Tan, and J.~Zhang, ``Sr-lda: Mining effective
  representations for generating service ecosystem knowledge maps,'' in
  \emph{2017 IEEE International Conference on Services Computing (SCC)}.\hskip
  1em plus 0.5em minus 0.4em\relax IEEE, 2017, pp. 124--131.

\bibitem{2014Generative}
I.~J. Goodfellow, J.~Pouget-Abadie, M.~Mirza, X.~Bing, and Y.~Bengio,
  ``Generative adversarial nets,'' \emph{MIT Press}, 2014.

\bibitem{he2016keyword}
Q.~He, R.~Zhou, X.~Zhang, Y.~Wang, D.~Ye, F.~Chen, J.~C. Grundy, and Y.~Yang,
  ``Keyword search for building service-based systems,'' \emph{IEEE
  Transactions on Software Engineering}, vol.~43, no.~7, pp. 658--674, 2016.

\bibitem{he2017efficient}
Q.~He, R.~Zhou, X.~Zhang, Y.~Wang, D.~Ye, F.~Chen, S.~Chen, J.~Grundy, and
  Y.~Yang, ``Efficient keyword search for building service-based systems based
  on dynamic programming,'' in \emph{International Conference on
  Service-Oriented Computing}.\hskip 1em plus 0.5em minus 0.4em\relax Springer,
  2017, pp. 462--470.

\bibitem{xia2014category}
B.~Xia, Y.~Fan, W.~Tan, K.~Huang, J.~Zhang, and C.~Wu, ``Category-aware api
  clustering and distributed recommendation for automatic mashup creation,''
  \emph{IEEE Transactions on Services Computing}, vol.~8, no.~5, pp. 674--687,
  2014.

\bibitem{al2015semantic}
M.~Al-Hassan, H.~Lu, and J.~Lu, ``A semantic enhanced hybrid recommendation
  approach: A case study of e-government tourism service recommendation
  system,'' \emph{Decision Support Systems}, vol.~72, pp. 97--109, 2015.

\bibitem{karthikeyan2018fuzzy}
N.~Karthikeyan, R.~K. RS \emph{et~al.}, ``Fuzzy service conceptual ontology
  system for cloud service recommendation,'' \emph{Computers \& Electrical
  Engineering}, vol.~69, pp. 435--446, 2018.

\bibitem{rupasingha2019alleviating}
R.~A. Rupasingha and I.~Paik, ``Alleviating sparsity by specificity-aware
  ontology-based clustering for improving web service recommendation,''
  \emph{IEEJ Transactions on Electrical and Electronic Engineering}, vol.~14,
  no.~10, pp. 1507--1517, 2019.

\bibitem{li2014novel}
C.~Li, R.~Zhang, J.~Huai, and H.~Sun, ``A novel approach for api recommendation
  in mashup development,'' in \emph{2014 IEEE International Conference on Web
  Services}.\hskip 1em plus 0.5em minus 0.4em\relax IEEE, 2014, pp. 289--296.

\bibitem{zhong2016web}
Y.~Zhong, Y.~Fan, W.~Tan, and J.~Zhang, ``Web service recommendation with
  reconstructed profile from mashup descriptions,'' \emph{IEEE Transactions on
  Automation Science and Engineering}, vol.~15, no.~2, pp. 468--478, 2016.

\bibitem{rosen2012author}
M.~Rosen-Zvi, T.~Griffiths, M.~Steyvers, and P.~Smyth, ``The author-topic model
  for authors and documents,'' \emph{arXiv preprint arXiv:1207.4169}, 2012.

\bibitem{bai2017dltsr}
B.~Bai, Y.~Fan, W.~Tan, and J.~Zhang, ``Dltsr: A deep learning framework for
  recommendations of long-tail web services,'' \emph{IEEE Transactions on
  Services Computing}, vol.~13, no.~1, pp. 73--85, 2017.

\bibitem{wang2019characterizing}
Z.~Wang, Z.~Dai, B.~P{\'o}czos, and J.~Carbonell, ``Characterizing and avoiding
  negative transfer,'' in \emph{Proceedings of the IEEE/CVF Conference on
  Computer Vision and Pattern Recognition}, 2019, pp. 11\,293--11\,302.

\bibitem{wang2019duskg}
H.~Wang, Z.~Wang, S.~Hu, X.~Xu, S.~Chen, and Z.~Tu, ``Duskg: A fine-grained
  knowledge graph for effective personalized service recommendation,''
  \emph{Future Generation Computer Systems}, vol. 100, pp. 600--617, 2019.

\bibitem{mezni2021context}
H.~Mezni, D.~Benslimane, and L.~Bellatreche, ``Context-aware service
  recommendation based on knowledge graph embedding,'' \emph{IEEE Transactions
  on Knowledge and Data Engineering}, 2021.

\bibitem{zheng2012collaborative}
Z.~Zheng, H.~Ma, M.~R. Lyu, and I.~King, ``Collaborative web service qos
  prediction via neighborhood integrated matrix factorization,'' \emph{IEEE
  Transactions on Services Computing}, vol.~6, no.~3, pp. 289--299, 2012.

\bibitem{chen2010regionknn}
X.~Chen, X.~Liu, Z.~Huang, and H.~Sun, ``Regionknn: A scalable hybrid
  collaborative filtering algorithm for personalized web service
  recommendation,'' in \emph{2010 IEEE international conference on web
  services}.\hskip 1em plus 0.5em minus 0.4em\relax IEEE, 2010, pp. 9--16.

\bibitem{maaradji2011social}
A.~Maaradji, H.~Hacid, R.~Skraba, and A.~Vakali, ``Social web mashups full
  completion via frequent sequence mining,'' in \emph{2011 IEEE World Congress
  on Services}.\hskip 1em plus 0.5em minus 0.4em\relax IEEE, 2011, pp. 9--16.

\bibitem{qi2017data}
L.~Qi, Z.~Zhou, J.~Yu, and Q.~Liu, ``Data-sparsity tolerant web service
  recommendation approach based on improved collaborative filtering,''
  \emph{IEICE TRANSACTIONS on Information and Systems}, vol. 100, no.~9, pp.
  2092--2099, 2017.

\bibitem{xie2019integrated}
F.~Xie, J.~Wang, R.~Xiong, N.~Zhang, Y.~Ma, and K.~He, ``An integrated service
  recommendation approach for service-based system development,'' \emph{Expert
  Systems With Applications}, vol. 123, pp. 178--194, 2019.

\bibitem{liang2016meta}
T.~Liang, L.~Chen, J.~Wu, H.~Dong, and A.~Bouguettaya, ``Meta-path based
  service recommendation in heterogeneous information networks,'' in
  \emph{International Conference on Service-Oriented Computing}.\hskip 1em plus
  0.5em minus 0.4em\relax Springer, 2016, pp. 371--386.

\bibitem{jain2015aggregating}
A.~Jain, X.~Liu, and Q.~Yu, ``Aggregating functionality, use history, and
  popularity of apis to recommend mashup creation,'' in \emph{International
  Conference on Service-Oriented Computing}.\hskip 1em plus 0.5em minus
  0.4em\relax Springer, 2015, pp. 188--202.

\bibitem{samanta2017recommending}
P.~Samanta and X.~Liu, ``Recommending services for new mashups through service
  factors and top-k neighbors,'' in \emph{2017 IEEE International Conference on
  Web Services (ICWS)}, 2017, pp. 381--388.

\bibitem{xiong2018deep}
R.~Xiong, J.~Wang, N.~Zhang, and Y.~Ma, ``Deep hybrid collaborative filtering
  for web service recommendation,'' \emph{Expert systems with Applications},
  vol. 110, pp. 191--205, 2018.

\bibitem{chen2018software}
L.~Chen, A.~Zheng, Y.~Feng, F.~Xie, and Z.~Zheng, ``Software service
  recommendation base on collaborative filtering neural network model,'' in
  \emph{International Conference on Service-Oriented Computing}.\hskip 1em plus
  0.5em minus 0.4em\relax Springer, 2018, pp. 388--403.

\bibitem{he2017neural}
X.~He, L.~Liao, H.~Zhang, L.~Nie, X.~Hu, and T.-S. Chua, ``Neural collaborative
  filtering,'' in \emph{Proceedings of the 26th international conference on
  world wide web}, 2017, pp. 173--182.

\bibitem{9492754}
H.~Wu, Y.~Duan, K.~Yue, and L.~Zhang, ``Mashup-oriented web api recommendation
  via multi-model fusion and multi-task learning,'' \emph{IEEE Transactions on
  Services Computing}, pp. 1--1, 2021.

\bibitem{2016Affine}
I.~E. Rube, ``Affine transformation,'' \emph{Introduction to Geometric
  Computing}, vol.~94, no.~3, pp. 473--481, 2016.

\bibitem{trivedi2019dyrep}
\emph{Dyrep: Learning representations over dynamic graphs}, vol. International
  Conference on Learning Representations.\hskip 1em plus 0.5em minus
  0.4em\relax OpenReview.net, 2019.

\bibitem{kingma2014adam}
\BIBentryALTinterwordspacing
D.~P. Kingma and J.~Ba, ``Adam: {A} method for stochastic optimization,'' in
  \emph{3rd International Conference on Learning Representations, {ICLR} 2015,
  San Diego, CA, USA, May 7-9, 2015, Conference Track Proceedings}, Y.~Bengio
  and Y.~LeCun, Eds., 2015. [Online]. Available:
  \url{http://arxiv.org/abs/1412.6980}
\BIBentrySTDinterwordspacing

\bibitem{wolf-etal-2020-transformers}
T.~Wolf, L.~Debut, V.~Sanh, J.~Chaumond, C.~Delangue, A.~Moi, P.~Cistac,
  T.~Rault, R.~Louf, M.~Funtowicz, J.~Davison, S.~Shleifer, P.~von Platen,
  C.~Ma, Y.~Jernite, J.~Plu, C.~Xu, T.~L. Scao, S.~Gugger, M.~Drame, Q.~Lhoest,
  and A.~M. Rush, ``Transformers: State-of-the-art natural language
  processing,'' in \emph{2020 Conference on Empirical Methods in Natural
  Language Processing: System Demonstrations}.\hskip 1em plus 0.5em minus
  0.4em\relax ACL, 2020, pp. 38--45.

\bibitem{mikolov2013word2vec}
T.~Mikolov, K.~Chen, G.~Corrado, and J.~Dean, ``Efficient estimation of word
  representations in vector space,'' 2013.

\bibitem{mcinnes2018umap-software}
L.~McInnes, J.~Healy, N.~Saul, and L.~Grossberger, ``Umap: Uniform manifold
  approximation and projection,'' \emph{The Journal of Open Source Software},
  vol.~3, no.~29, p. 861, 2018.

\end{thebibliography}
%

%



\begin{IEEEbiography}[{\includegraphics[width=1in,height=1.25in,clip,keepaspectratio]{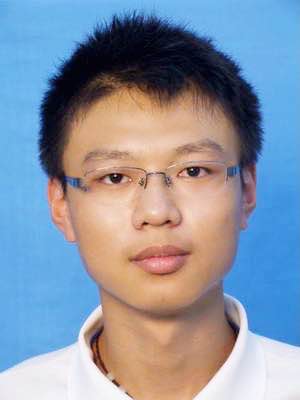}}]{Mingyi Liu}
received his B.S. degree from the School of Computer Science and Technology, Harbin Institute of Technology in 2018. He is currently pursuing the Ph.D. degree in software engineering at Harbin Institute of Technology (HIT), China. His research interests include service ecosystem model, service evolution analysis, data mining and knowledge graph.
\end{IEEEbiography}
\begin{IEEEbiography}[{\includegraphics[width=1in,height=1.25in,clip,keepaspectratio]{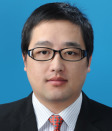}}]{Zhiying Tu}
 is an associate professor of the School of Computer Science and Technology at Harbin Institute of Technology (HIT). He holds a PhD degree in Computer Integrated Manufacturing (Productique) from the University of Bordeaux. Since 2013, he start to work at HIT. His research interest are Service Computing, Enterprise Interoperability, and Cognitive Computing. He has 20 publications as edited books and proceedings, refereed book chapters, and refereed technical papers in journals and conferences. He is the member of IEEE Computer Society, and CCF China.
\end{IEEEbiography}

\begin{IEEEbiography}[{\includegraphics[width=1in,height=1.25in,clip,keepaspectratio]{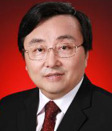}}]{Xiaofei Xu}
is a professor or at School of Computer Science and Technology, and Vice President of Harbin Institute of Technology. He received the Ph.D. degree in computer science from Harbin Institute of Technology in 1988. His research interests include enterprise intelligent computing, services computing, Internet of services, and data mining. He is the Associate Chair of IFIP TC5 WG5.8, chair of INTEROP-VLab China Pole, Fellow of China Computer Federation (CCF), and the vice director of the technical committee of service computing of CCF. He is the author of more than 300 publications. He is member of the IEEE and ACM.
\end{IEEEbiography}

\begin{IEEEbiography}[{\includegraphics[width=1in,height=1.25in,clip,keepaspectratio]{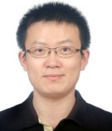}}]{Zhongjie Wang}
 is a professor at School of Computer Science and Technology, Harbin Institute of Technology (HIT). He received the Ph.D. degree in computer science from Harbin Institute of Technology in 2006. His research interests include services computing, mobile and social networking services, and software architecture. He is the author of more than 40 publications. He is a member of the IEEE.
\end{IEEEbiography}




\end{document}